\documentclass[oneside,reqno]{amsart}
\usepackage[latin1]{inputenc}
\usepackage[english]{babel} 
\usepackage{ifpdf} 
\ifpdf
\usepackage[pdftex]{graphicx} 
\usepackage{epstopdf} 
\else
\usepackage[dvips]{graphicx}  
\fi 
\usepackage{caption} 
\usepackage[pdftitle={Supervised FXI classification},
  pdfauthor={Liu, Schot, Engblom},
  pdffitwindow=true,
  breaklinks=true,
  colorlinks=true,
  urlcolor=blue,
  linkcolor=red,
  citecolor=red,
  anchorcolor=red]{hyperref}
\usepackage{amssymb}
\usepackage{amsmath}
\usepackage{amsthm}
\usepackage[flushleft]{threeparttable}

\usepackage{algorithm}
\usepackage[subrefformat=parens,labelformat=parens,labelfont={rm}]{subfig}
\usepackage{listings}
\usepackage{mathrsfs}	
\usepackage[numbers,sort]{natbib} 
\usepackage{amsmath,mathrsfs}
\usepackage[inline]{enumitem}

\usepackage{ragged2e}

\newcommand{\Mpix}{M_{\mbox{{\tiny pix}}}}
 
\newcommand{\Mdata}{M_{\mbox{{\tiny data}}}}

\begin{document}
\title[Supervised FXI classification]{Supervised Classification
  Methods for Flash X-ray single particle diffraction Imaging}

\author[J.~Liu]{Jing Liu}
\author[G.~Schot]{Gijs van der Schot}	
\author[S.~Engblom]{Stefan Engblom} 

\address[J.~Liu]{Laboratory of Molecular Biophysics, Department of
  Cell and Molecular Biology, Uppsala university, SE-751 24 Uppsala,
  Sweden.} \email{jing.liu@icm.uu.se}
  
\address[G.~Schot]{Cryo-Electron Microscopy, Bijvoet Center for
  Biomolecular Research, Utrecht University, 3584 CH Utrecht, The
  Netherlands.} \email{g.vanderschot@uu.nl}

\address[J.~Liu  \and S.~Engblom]{Division of Scientific Computing,
  Department of Information Technology, Uppsala university, SE-751 05
  Uppsala, Sweden.} \email{jing.liu, stefane@it.uu.se}

\thanks{Corresponding author: S.~Engblom, telephone +46-18-471 27 54,
  fax +46-18-51 19 25.}
\urladdr[S. Engblom]{\url{http://user.it.uu.se/~stefane}}
	
\date{\today}
\maketitle
	
%**************************************************************************
	
\selectlanguage{english}
	
\begin{abstract}
  Current Flash X-ray single-particle diffraction Imaging (FXI)
  experiments, which operate on modern X-ray Free Electron Lasers
  (XFELs), can record millions of interpretable diffraction patterns
  from individual biomolecules per day.  Due to the stochastic nature
  of the XFELs, those patterns will to a varying degree include
  scatterings from contaminated samples. Also, the heterogeneity of
  the sample biomolecules is unavoidable and complicates data
  processing. Reducing the data volumes and selecting high-quality
  single-molecule patterns are therefore critical steps in the
  experimental set-up.

  In this paper, we present two supervised template-based learning
  methods for classifying FXI patterns. Our Eigen-Image and
  Log-Likelihood classifier can find the best-matched template for a
  single-molecule pattern within a few milliseconds. It is also
  straightforward to parallelize them so as to fully match the XFEL
  repetition rate, thereby enabling processing at site.
\end{abstract}


%**************************************************************************

\section{Introduction}

Modern X-ray Free Electron Laser (XFEL) technology has provided the
opportunity for exploring biological structures from individual
biological particles, rather than relying on crystallization-based
technologies. It is therefore potentially possible to investigate
biomolecules or biological processes that are intrinsically
dynamic. XFELs produce X-ray pulses shorter than $50$ femtosecond
(fs), which are $10^9$ times more brilliant than the radiation
produced in conventional synchrotrons. The ultra-short and extremely
bright X-ray pulses outrun the radiation damage and allow the
recording of sufficiently strong and interpretable 2-dimensional (2D)
diffraction patterns from single biological particles
\cite{Neutze2000, Chapman2006}. This principle is called
diffract-and-destroy and has been shown to be successful for particles
as large as small cells, and down to viruses smaller than 50
nanometers (nm)~\cite{Seibert2011,Hantke2014,Schot2015,Daurer2017}.

Another feature of XFELs is their high repetition rates. The Linac
Coherent Light Source (LCLS)~\cite{LCLS} operates at 120 Hz and can
produce over 400,000 diffraction patterns per hour, i.e., more than
1.6 TB per hour or 38 TB per day. The massive volume of data makes
manual classification of diffraction patterns impractical. The
challenge is much more severe in the newest facility --- the European
XFEL~\cite{EXFEL}, which operates at up to 27,000 Hz and can store
more than 3 million images per hour. Ideally, all these images would
originate from one single biomolecule per exposure. However, the
detector also records diffracted signals from multiple scatterers such
as particle clusters, buffer impurities, and contaminant materials as
discussed in \cite{Hantke2014, Daurer2017}.

In order to assemble the 2D diffraction patterns into 3D structures,
it is essential that data frames are classified and that diffraction
patterns originating from contaminants and multiple molecules are
sorted out. In 2014, a real-time rejection method~\cite{autoselect}
was proposed to select diffraction patterns by thresholding and using
Time-of-Flight spectroscopy. Previous sorting algorithms was based on
support vector machines and on spectral clustering techniques
\cite{pcasorting,spcl}.

In this paper, we develop two template-based classification methods
for particle selection --- the Eigen-Image (EI) and the Log-Likelihood
(LL) method. Both methods assess the similarity between template
diffraction patterns and incoming patterns by analyzing eigenvector
projections and log-likelihood function, respectively. In
\S\ref{sec:FXI}, we briefly describe a typical Flash X-ray
single-particle diffraction Imaging (FXI) experiment. Next, we
introduce the EI and the LL method for classification in
\S\ref{sec:method}. Following data descriptions in \S\ref{sec:data},
we perform numerical experiments to evaluate the sharpness of our
classification methods in \S\ref{sec:experiment}. A concluding
discussion is found in \S\ref{sec:Conclusion}.

%**************************************************************************

\section{Flash X-ray single particle diffraction Imaging (FXI)}
\label{sec:FXI}

\begin{figure*}[!htb]
\centering
\subfloat[]{\includegraphics[width = 0.8\textwidth]{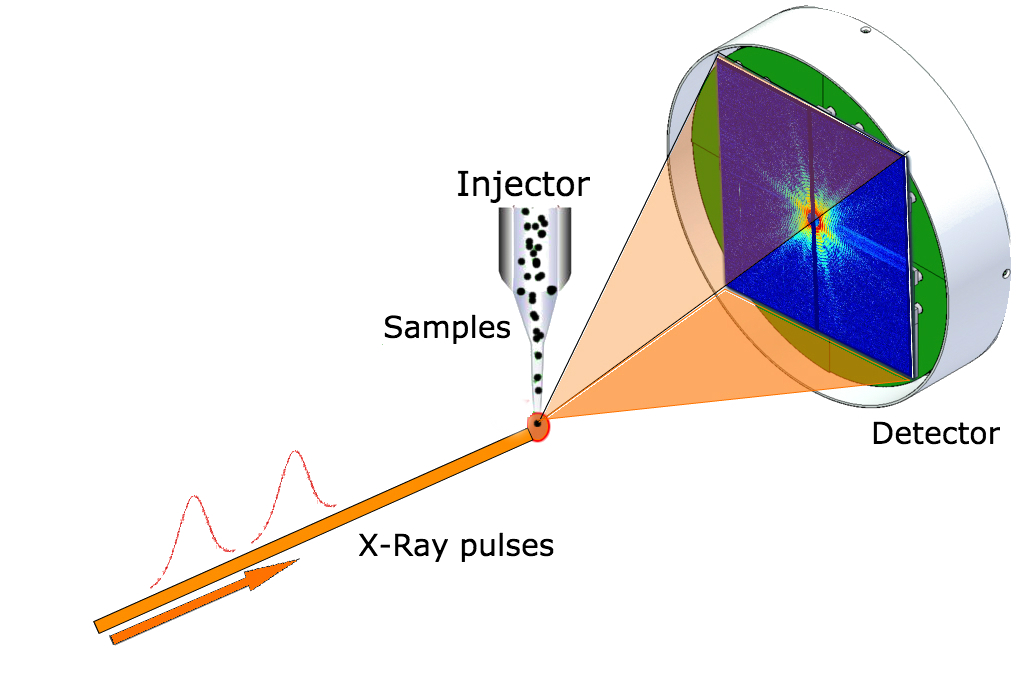}\label{fig:detector}}\\
\subfloat[]{\includegraphics[width = 0.23\textwidth]{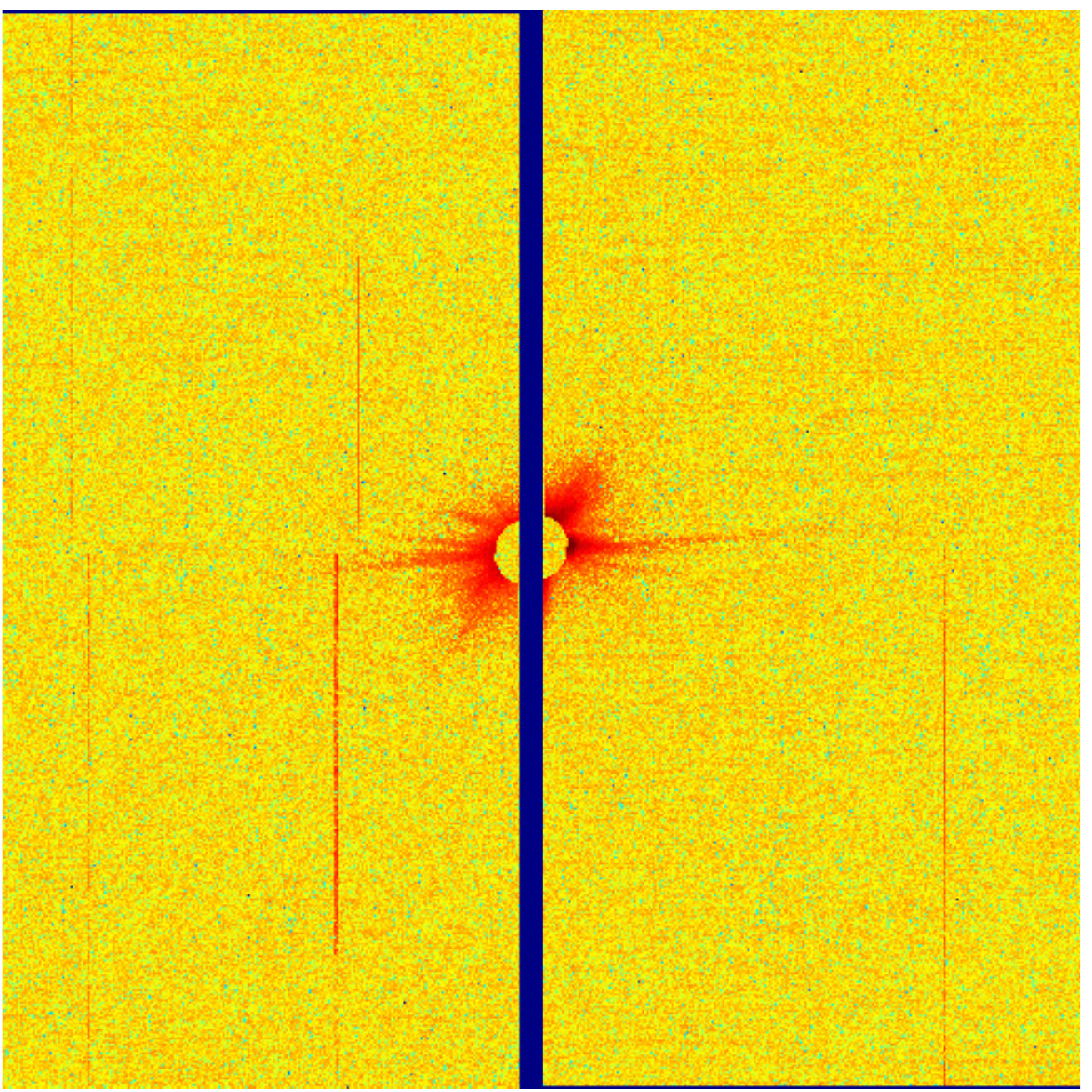} \label{fig:megBlank}}
\subfloat[]{\includegraphics[width = 0.23\textwidth]{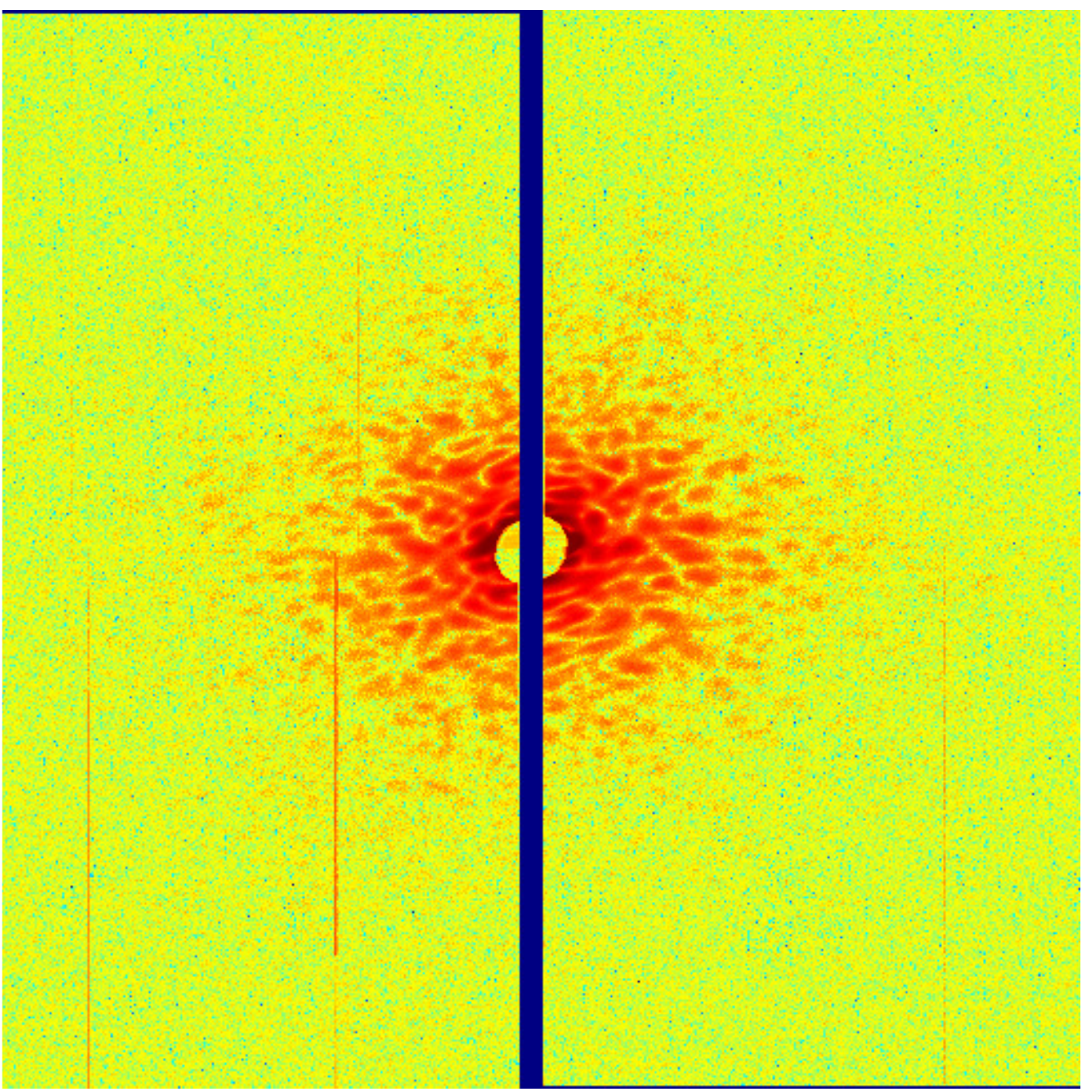} \label{fig:megMulti}}
\subfloat[]{\includegraphics[width = 0.23\textwidth]{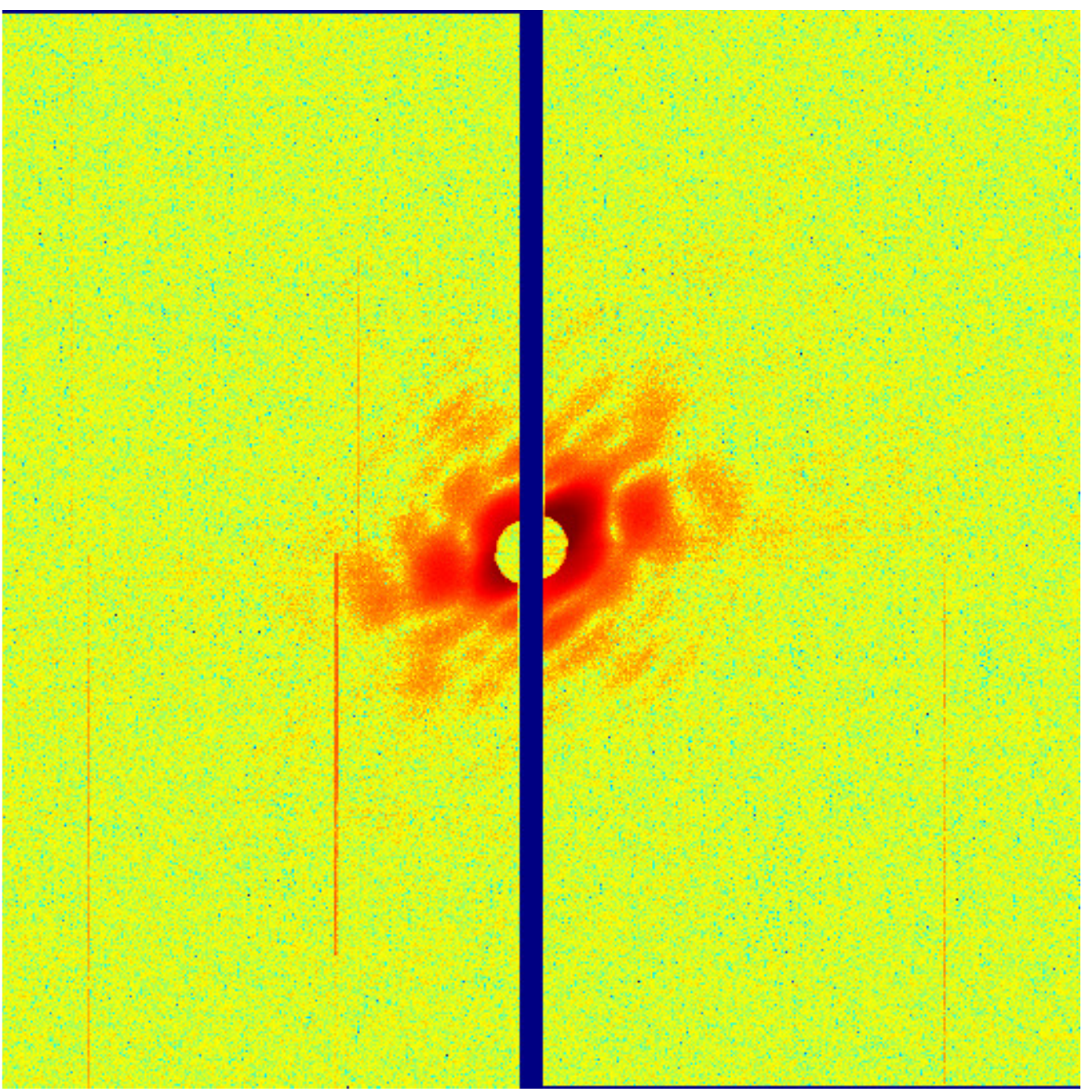} \label{fig:megCon}}
\subfloat[]{\includegraphics[width = 0.23\textwidth]{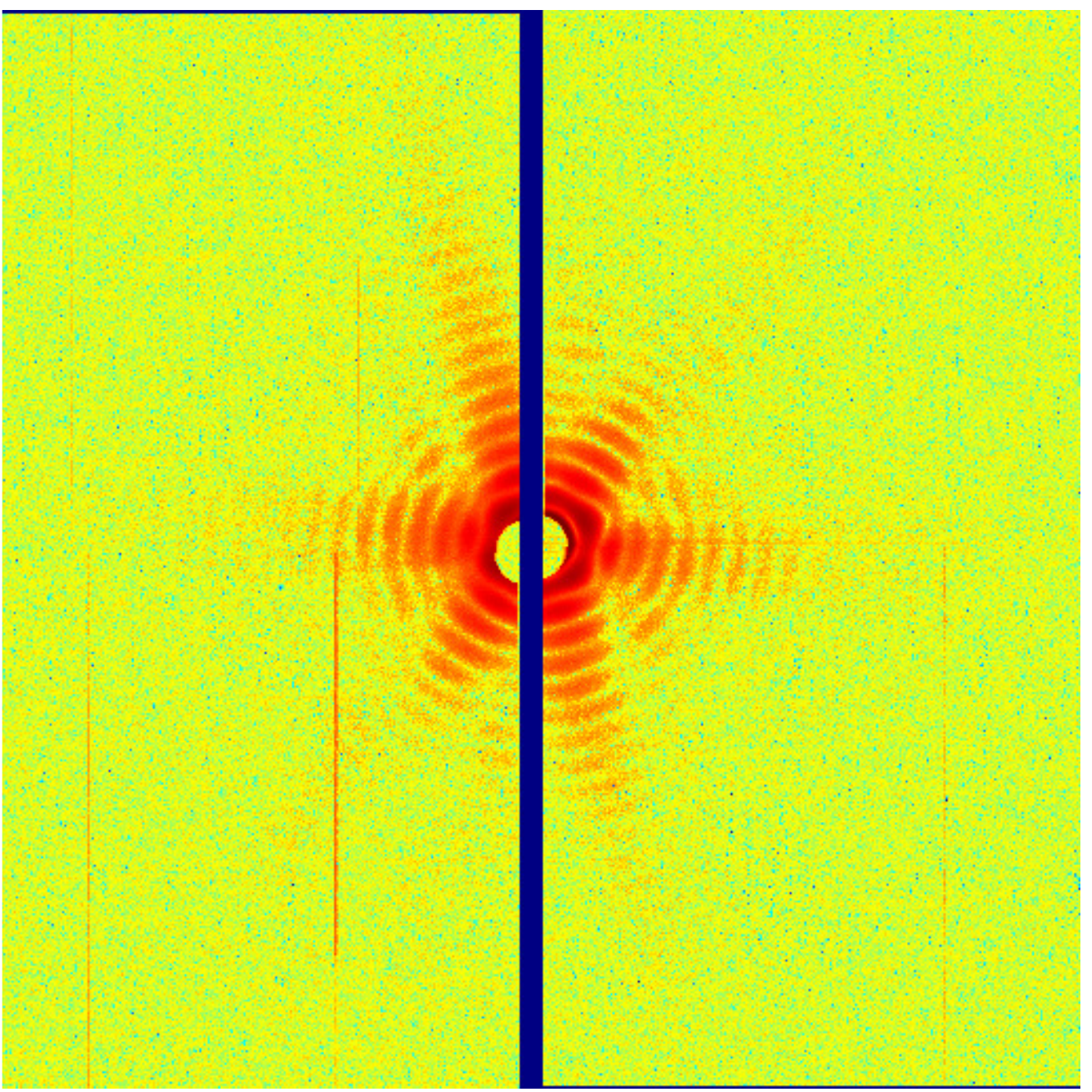} \label{fig:megSingle}}
\caption{\protect\subref{fig:detector}: A typical setup of an FXI
  experiment. [\protect\subref{fig:megBlank}--\protect\subref{fig:megSingle}]:
  Four frames captured in an FXI experiment.
  \protect\subref{fig:megBlank} was a blank frame which contains only
  background scattering. [\protect\subref{fig:megMulti} and
    \protect\subref{fig:megCon}] were frames from multiple particles
  or with contaminants.  \protect\subref{fig:megSingle} was a
  single-particle frame from an icosahedral-shape virus with a
  relatively strong signal. This is the most interesting pattern and
  can be used for assembling 3D structure in later steps. All
  diffraction patterns are displayed in logarithmic scale.}
\label{fig:setup}
\end{figure*}

For a typical FXI experiment, the diffraction data acquired is
depicted graphically in Figure~\ref{fig:setup}. A stream of biological
molecules is injected into the X-ray interaction region, where sample
particles interact with incoming coherent X-ray pulses, resulting in a
collection of diffraction patterns on the detector. This procedure is
a stochastic process as the interactions between particles and X-ray
pulses occur at random. Firstly, the number of particles at the
interaction point is unobserved, i.e., we may obtain blank frames with
only background noises, single-particle patterns, multiple-particles
patterns, and frames with signals from contaminants. Secondly, the
current FXI technology cannot monitor the orientations of particles,
and therefore extra steps are necessary to recover the 3D structure
from single-particle frames. Last but not least, the strengths of the
diffraction signals vary a lot, mainly due to the stochastic nature of
the XFELs and the different locations of particles in the interaction
region, respectively. The relative strength of the diffraction signal
is referred to as photon fluence, and we denote it by $\phi$.

Typical FXI setups use discrete digital detectors, and therefore the
captured frames are also discrete. Further, some pixel counts near the
center are inaccessible or overflow as a result of physical
limitations and arrangements of the detector.

%**************************************************************************

\section{Classification Methods}
\label{sec:method}

Template-based methods for classifying diffraction patterns allow
identifying the class of an unlabeled diffraction pattern by searching
for its best-matched template. For such methods, the collection of
templates is referred to as the \emph{training dataset}, an unlabeled
pattern is called a \emph{testing image}, and the classification
procedure is referred to as the \emph{classifier}. In this section, we
discuss two classifiers --- the Eigen-Image
(EI)~\cite{eigeneyes,eigenlights,pcaEigen, AlexPentland,
  Karhunen-Loeve_procedure, eigenface} and the Log-Likelihood (LL)
classifier~\cite{ll1,ll2,ll3}, to classify a testing diffraction
pattern relying on the training dataset.

\subsection{Eigen-Image (EI) Classifier}
\label{subsec:method_PCA}

The EI method has two steps --- the training and the classification
step.  In the training step, we train our EI classifier by projecting
the training dataset to its eigenvectors. In the classification step,
we label a testing image by minimizing the distance between the
eigenvector projections of the testing image and the training dataset.

Let i.i.d.~template diffraction patterns $T = (T_k)_{k=1}^{\Mdata}$ be
the training dataset, consisting of $\Mdata$ frames. Since the
detector is discrete, we denote the $k$th pattern by $T_k =
(T_{ik})_{i=1}^{\Mpix}$. To train an EI classifier, we first transfer
the training dataset $T$ into the image space $A$ by the shift
\begin{align}
  A = (A_k)_{k=1}^{\Mdata} = (T_k - \bar T)_k,
\end{align}
where $\bar T$ is the pixel average of the training dataset, 
\begin{align}
  \bar T = \frac{1}{\Mdata} \sum_{k=1}^{\Mdata} T_k.
\end{align}

Practically, the covariance matrix of $A$ ($AA^T$) is too large to
decompose into eigenvectors. Therefore we factorize the matrix $A^TA$
by instead,
\begin{align}
  A^TA = V \Lambda V^T,
\end{align}
where $\Lambda$ is the main diagonal matrix, whose diagonal elements
are the corresponding eigenvalues, and $V$ is the matrix of
eigenvectors of $A^TA$. We can now compute the eigenvectors of the
covariance matrix $AA^T$ by
\begin{align}
  U =  A V,
\end{align}
and $U$ is sometimes also referred to as
eigenfaces~\cite{eigeneyes,eigenface}.

The eigenvector projection matrix of the image space $A$ is defined as
follows:
\begin{align}
\Omega = U^T A.
\end{align}

Using $U$ and $\Omega$, we can now classify a testing diffraction
pattern $P = (P_i)_{i=1}^{\Mpix}$, by minimizing the euclidean
distance between its eigenvector projection matrix $W$ and $\Omega$,
\begin{align}
\arg_k \min ||W_k - \Omega_k ||_{L_2},
\label{eq:optei1}
\end{align}
where
\begin{align}
W = U^T(P-\bar T).
\label{eq:optei2}
\end{align}

\subsection{Log-Likelihood (LL) Classifier}

The LL Classifier attempts to classify a testing image by maximizing
the log-likelihood function of a given probability density
function. Since the photon counting procedure is assumed to obey the
Poisson distribution, we can write the joint likelihood function as
follows:
\begin{align}
  \prod_{i=1}^{\Mpix} \textbf{P}(P_i| T_{ik},\phi_k) =   \prod_{i=1}^{\Mpix} \dfrac{(\phi_k T_{ik})^{P_i} e^{-\phi_k T_{ik}}}{P_i!} =: \mathcal{Q}_{ik},
  \label{eq:Qik}
\end{align}
where $\phi$ is the photon fluence (relative signal strength), and can
be estimated by
\begin{align}
  \phi_k = \dfrac{\sum_{i=1}^{\Mpix} P_i}{ \sum_{i=1}^{\Mpix} T_{ik}}.
  \label{eq:phi_all}
\end{align} 

The joint log-likelihood function $\mathcal{L}$ for the LL classifier
is therefore
\begin{align}
  \log(\mathcal{Q}_{ik}) \propto \sum_{i=1}^{\Mpix} P_i \log(T_{ik}) + P_i \log(\phi_k) -\phi_k T_{ik} =: \mathcal{L}_k,
  \label{eq:ll}
\end{align} 

We can now classify the testing image $P$ by simply maximizing the
joint log-likelihood function in \eqref{eq:ll}:
\begin{align}
  \arg_k \max \mathcal{L}_k = \arg_k \max \sum_{i=1}^{\Mpix} P_i \log(T_{ik}) + P_i\log(\phi_k) -\phi_k T_{ik}.
  \label{eq:optll}
\end{align}

For classifying multiple testing images, the EI method computes $U$
and $\Omega$ only once, and hence less computations are needed
compared to the LL method.

%**************************************************************************

\section{Data Description}
\label{sec:data}

In this section we describe our training and testing datasets. Since
most viruses have either helical or icosahedral capsid structure
\cite{capsid1,capsid2}, we used regular uniform-density icosahedrons
to generate diffraction patterns via Condor~\cite{Condor}. For our
simulations, we used a setup similar to the beam profile of the FXI
mimivirus experiment \cite{mimivirus_Xrays}. More specifically, we
used $1$ $nm$ X-ray pulses, with a peak energy of $1$ $mJ$. We also
assumed that the X-ray pulses had a circular focus of $10$ $\mu m$ in
diameter. Further, the distance between the detector and the
interaction region was $0.74$ meters, and the detector itself was
$960\times 960$ pixels, or $72\times 72$ $\mu m^2$. Finally, a
circular missing-data area of $80$ pixels in diameter was set to zero.

\begin{figure}[!htb]
  \centering
  \subfloat[]{\includegraphics[width=.32\textwidth]{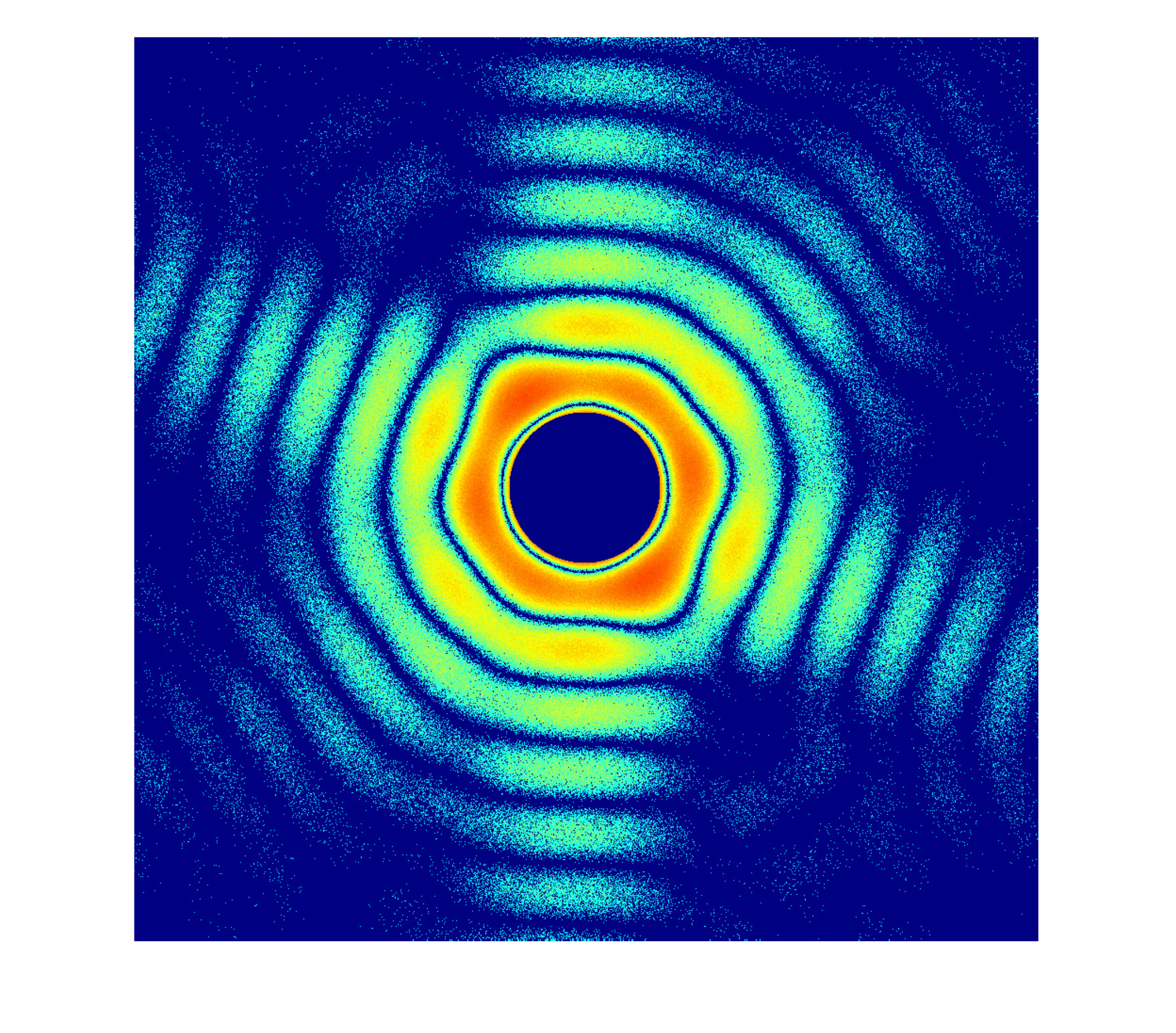}\label{fig:S_s_180}}
  \subfloat[]{\includegraphics[width=.32\textwidth]{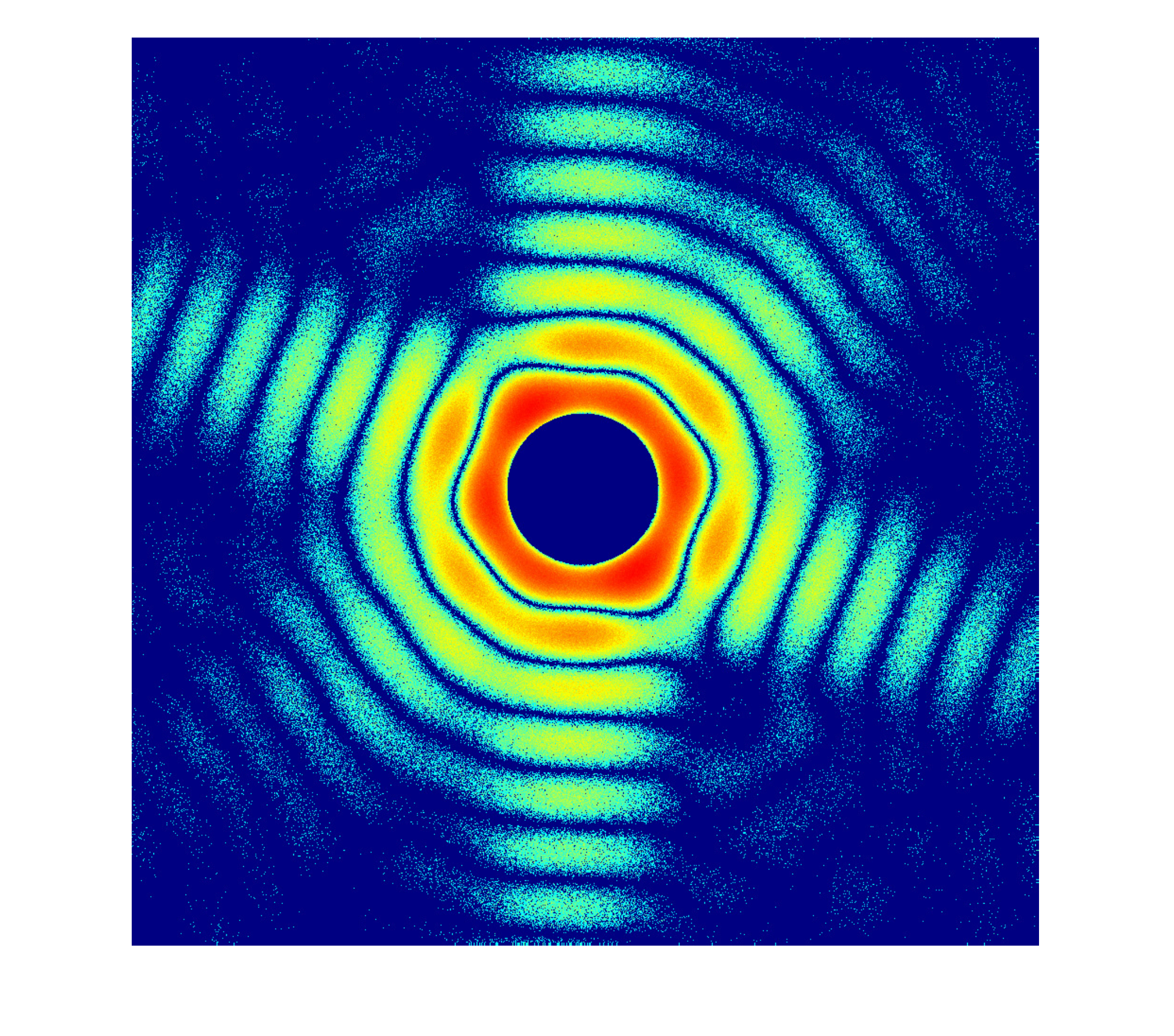}\label{fig:S_s_200}}
  \subfloat[]{\includegraphics[width=.335\textwidth]{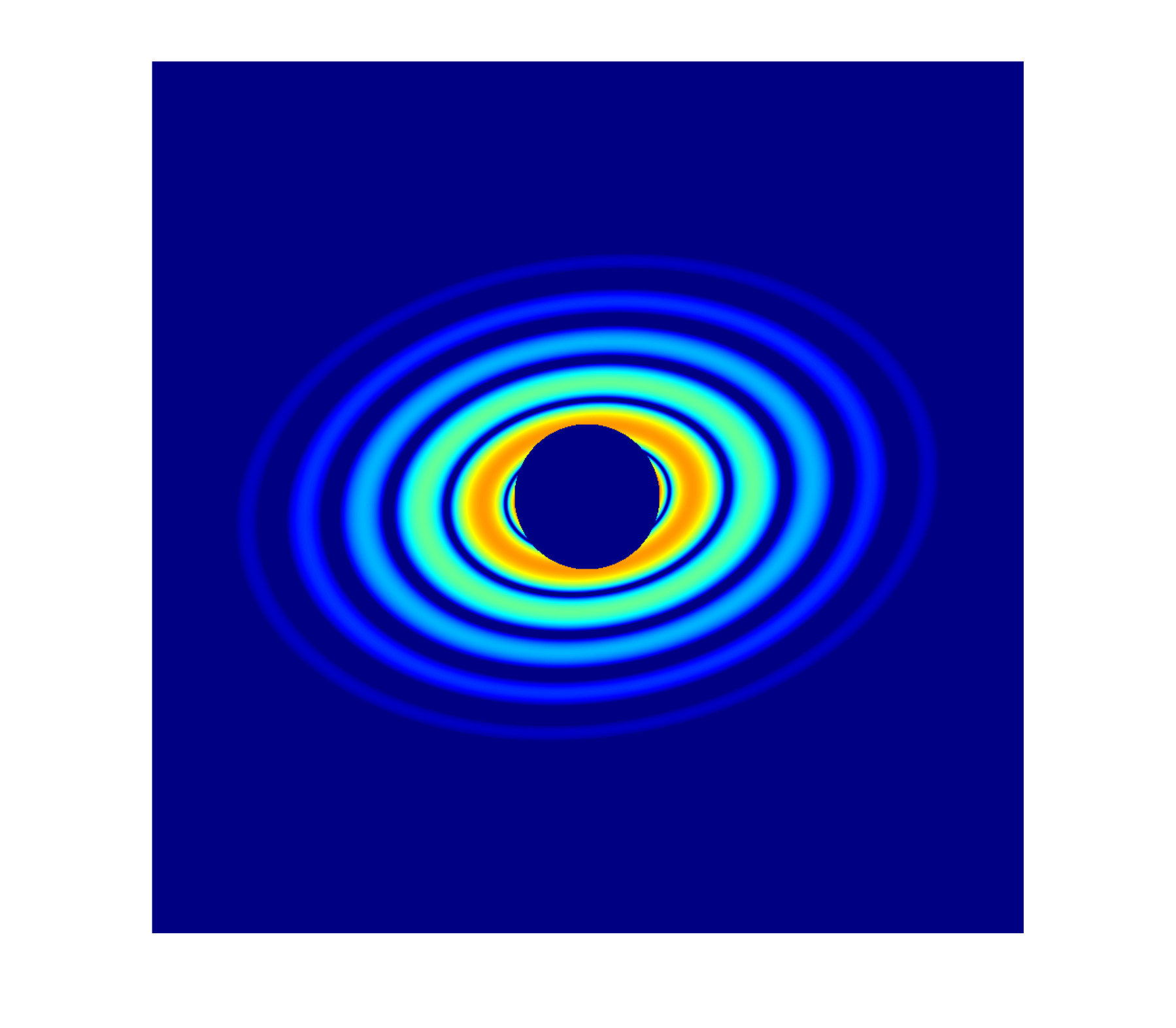}\label{fig:S_s_sph}}\\
  \caption{ \protect\subref{fig:S_s_180}: A noisy diffraction pattern
    from a $180$ $nm$ icosahedron. \protect\subref{fig:S_s_200}: A
    noisy pattern from a $200$ $nm$ icosahedron in the same particle
    orientation.  \protect\subref{fig:S_s_sph}: A noiseless pattern
    from a $180$ $nm$ spheroid.}
\label{fig:S_s}
\label{fig:ty_patterns}
\end{figure}

To assess our classifiers systematically, we gradually increased the
complexity of the testing dataset. With five synthetic testing
datasets, we mimicked diffraction patterns of particles with noise,
different fluences, and of various sizes and shapes. We also evaluated
our methods for the actual mimivirus FXI data \cite{3d_mimi}.
Figure~\ref{fig:ty_patterns} illustrates two noisy icosahedral
diffraction patterns at particle sizes $180$ $nm$ and $200$ $nm$ in
the same particle orientation, and one spheroid diffraction pattern at
size $180$ $nm$.

\subsection{Homogeneous Datasets}

We first simulated diffraction patterns from a regular icosahedron of
$180$ $nm$ in diameter. The training dataset $T$ had $290$ frames, and
the Euclidean distances between two arbitrary patterns were larger or
equal than $220$. The first testing dataset $D$ was a noiseless
homogeneous dataset, which contained $\Mdata = 1000$ noiseless
icosahedral diffraction patterns. The first $290$ frames were from the
training dataset $T$ and were used as benchmarks. The rest $810$
frames were random-orientation patterns from the same icosahedron.

Since the photon counting procedure is assumed to follow the Poisson
distribution, we added Poissonian noise to $D$ for our noisy dataset
$P$,
\begin{align}
  P_k \sim Poisson(D_k), k = 1,2,\cdots,\Mdata.
  \label{eq:Tn}
\end{align}

By scaling $P$ with different fluences, we obtained our last
homogeneous testing dataset --- the scaled noisy dataset $F$ by
\begin{align}
  F_k \sim Poisson (\Phi_k D_k),
  \label{eq:Tnf}
\end{align}
where $\Phi_k$ was uniformly and randomly chosen between 0.01 to 1.1,
\begin{align}
  \label{eq:random_fluence}
  \Phi_k \sim \mathcal{U}\{0.01, 1.1\}.
\end{align} 

\subsection{Heterogeneous Particle Sizes}

Considering the potential size variation of viruses, we generated our
testing dataset $S$ ($\Mdata=2000$) from uniform-density icosahedrons
with randomly and uniformly chosen diameters between $150$ $nm$ and
$210$ $nm$ ($\sim \mathcal{U}\{150, 210\}$). Similar to $F$, all
patterns in $S$ were Poissonian with random fluences according to
\eqref{eq:random_fluence}.

\subsection{Heterogeneous Particle Shapes}

To mimic heterogeneous particle shapes, the synthetic testing dataset
$X$ contained diffraction patterns from both icosahedrons and
spheroids. The diameters of the objects varied from $150nm$ to
$210nm$, with changing fluences $\Phi_k \sim \mathcal{U}\{0.01,
1.1\}$. Further, the shapes of the spheroids were also changing, as
the aspect ratios of the spheroids (the ratio of the length of the
minor axis to the length of the major one) were varying between 0.6
and 1. In total, the dataset $X$ contained $\Mdata = 1200$ frames ---
200 spheroidal patterns and 1000 icosahedral patterns randomly
selected from $S$.

\subsection{Mimivirus Dataset}

To be relevant to real FXI experiments, we also classified the
mimivirus dataset~\cite{3d_mimi,mimivirus_Xrays}. To classify this
dataset, we generated a new training dataset with the corresponding
experimental beam profile~\cite{mimivirus_Xrays}, and the training
dataset contained $1000$ random-orientation frames of a $490$ $nm$
icosahedron.

To summarize, Table~\ref{tab:datasets} lists the primary parameters of
all datasets.

\begin{threeparttable}[!htbp]
  \centering
  \caption{Primary parameters of all datasets.}
  \begin{tabular}{l c c l c c}\hline
    Dataset			&  Diameter (nm)	& \# Patterns ($\Mdata$)	& Noise		& Fluence $\Psi$	 \\ \hline
    $T$\tnote{a}            & 180				& 290			& N/A		& 1 \\ 
    $D$\tnote{b}			& 180				& 1000			& N/A		& 1 \\ 
    $P$\tnote{b}   			& 180				& 1000			& Poisson	& 1 \\ 
    $F$\tnote{b}			& 180				& 1000 			& Poisson	& $\mathcal{U}\{0.01, 1.1\}$\tnote{c} \\ 
    $S$\tnote{b}			& $\mathcal{U}\{150, 210\}$\tnote{c} 			& 2000			& Poisson 	& $\mathcal{U}\{0.01, 1.1\}$\tnote{c} 	 \\ 
    $X$\tnote{d}  	    & $\mathcal{U}\{150, 210\}$\tnote{c} 		    & 1200			& Poisson	& $\mathcal{U}\{0.01, 1.1\}$\tnote{c} \\
    Mimivirus\tnote{e}	& $\approx 490$		& 198			& N/A    	& N/A \\
    \hline
  \end{tabular}
  \begin{tablenotes}
  \item[a] The training dataset for classifying synthetic testing datasets. 
  \item[b] Only synthetic regular icosahedral patterns were included in
    these datasets.
  \item[c] $\mathcal{U}$ is the uniform distribution.
  \item[d] $X$ contained 1000 random icosahedral frames from $S$ and
    $200$ spheroidal patterns with aspect ratio $\sim \mathcal{U}\{0.6,
    1\}$.
  \item[e] Patterns were as used in~\cite{3d_mimi,mimivirus_Xrays} and
    of icosahedral shape.
  \end{tablenotes}
  \label{tab:datasets}
\end{threeparttable}
 
%**************************************************************************

\section{Experiments}
\label{sec:experiment}

We now perform numerical experiments to investigate the efficiency and
accuracy of our EI and LL classifiers. For saving memory and execution
time without losing much accuracy in the classification, only the
central $480\times 480$ pixels were used in computations, and they
were binned into $120 \times 120$ pixels, i.e., every $4 \times 4$
pixels were averaged into one pixel.

\subsection{Error Metrics}

To compare the classification results, we define the following error
metrics.  Let $\Gamma_k = (\Gamma_{ik})_{i=1}^{\Mpix}$ be the $k$th
frame of the testing dataset $\Gamma$. Let $R=(R_i)_{i=1}^{\Mpix}$ be
the best-matched pattern of $\Gamma_k$ from the training dataset. The
classification error of $\Gamma_k$ with respect to $R$ is now defined
as follows:

\begin{align}
C_k (\Gamma_k,R) = \arg\min_{s,\hat \Phi_k}  \dfrac{\sum_{i=1}^{\Mpix}(\hat \Phi_k U_{ik}(R,s)  - \Gamma_{ik})^2}{\sum_{i=1}^{\Mpix} (\hat \Phi_k U_{ik}(R,s))^2},
\label{eq:L2_S}
\end{align}
where $\hat \Phi_k$ is the estimated fluence, 
\begin{align}
\hat \Phi_k = \dfrac{\sum_{i=1}^{\Mpix} \Gamma_{ik}}{\sum_{i=1}^{\Mpix}  U_{ik}(R,s)}.
\label{eq:phi_S}
\end{align} 
Further, $U_k(R,s)$ is an interpolation (or extrapolation) method that
resizes the pattern $R$ $s$ times and returns a scaled image at the
same size as $\Gamma_k$. Note that, $U(R,s) = R$ for our homogeneous
testing datasets $D$, $P$ and $F$, and $\hat \Phi_k = 1$ for the first
two.

Similarly, we define the fluence error by 
\begin{align}
E_k = \dfrac{ (\Phi_k -\hat \Phi_k)^2}{\Phi_k^2},
\label{eq:L2_phi_error}
\end{align} 
where $\Phi_k$ is the true fluence used to generate $\Gamma_k$.

%\begin{align}
%C_k(\Gamma_k,R) = \dfrac{\sum_{i=1}^{\Mpix}{ (\hat \phi_k  R_i - \Gamma_{ik})^2}}{\sum_{i=1}^{\Mpix}{ (\hat \phi_k R_i)^2}},
%\label{eq:l2-relative-error}
%\end{align} 
%where 
%\begin{align}
%\hat \phi_k = \dfrac{\sum_{i=1}^{\Mpix} \Gamma_{ik}}{\sum_{i=1}^{\Mpix}  R_i},
%\end{align}
%is the estimated fluence of $\Gamma_k$ with respect to $R$. This
%fluence estimation is needed only for the testing dataset $F$, and we
%set $\hat \phi_u = \phi = 1$ $1$ for the testing dataset $D$ and $P$.
%Similarly, we define the relative $L_2$ fluence error by
%\begin{align}
%E_k = \dfrac{ |(\phi_k -\hat \phi_k)|^2}{\phi_k^2},
%\label{eq:L2_phi_error}
%\end{align} 
%where $\phi_k$ is the fluence used to generate the testing pattern
%$\Gamma_k$.
%When particle size varies, the above error metric
%\eqref{eq:l2-relative-error} is unfair for any diffraction pattern
%whose size differs from the training dataset, hence we need first to
%fit the particle sizes before calculating classfication error.
%Again, let $\Gamma_k$ be $k$th frame of the testing dataset and $R$
%be its recognized pattern in the training dataset. An interpolation
%method $U_u(R,w)$ resizes $R$ by $w$ times and returns an image with
%the size of $\Gamma_u$.%, by padding zeros around if $w<1$ or cut
%border pixels if $w>1$.  To obtain the relative classification error
%together with the estimated fluence $\hat \phi_k$ and the resizing
%factor $w$, we solve the following optimization problem:

\subsection{Homogeneous Patterns}
\label{subsection: Odata}

We first tested our classifiers on the homogeneous synthetic datasets
--- the noiseless dataset $D$, the noisy dataset $P$, and the
scaled-Poisson dataset $F$, as listed in Table~\ref{tab:datasets}.

Since the first $290$ images of the three testing datasets were
modifications of the training dataset, we used them as benchmarks, and
compared their average classification error with the error from the
whole dataset. As listed in Table~\ref{tab:without_S}, we observed
that both classifiers matched all benchmark frames successfully with
classification errors around 0, 0.03 and 0.04 on datasets $D$, $P$ and
$F$, respectively. However, for the whole datasets, the EI classifier
performed slightly better than the LL classifier, obtaining about 1\%
less classification error.
 
\begin{table}[!htb]
  \centering
  \caption{The average classification error as defined in
    \eqref{eq:L2_S} of the EI and the LL classifier.  Benchmark is the
    average classification error of the benchmark patterns, and
    Complete denotes the error of the whole dataset.}
  \begin{tabular}{ l| c c| c c}
  \hline
  &  \multicolumn{2}{c|}{EI} &  \multicolumn{2}{c}{LL} \\ 
  Datset	&   Benchmark  &  Complete  	& Benchmark  		& Complete     \\ \hline
  $D$ 	&   0        & 0.035          & 0             &  0.041         \\ 
  $P$	&   0.031     & 0.051          & 0.031          &  0.062       \\ 
  $F$	&   0.042     & 0.053          & 0.043          &  0.064         \\ \hline
  \end{tabular}
  \label{tab:without_S}
\end{table}

The fluence error, as defined in \eqref{eq:L2_phi_error}, of the
dataset $F$ from the EI classifier was $0.035$, comparing with $0.049$
from the LL classifier. Further, the EI classifier was more efficient,
and took only $3.7$ $ms$ per image in a single-core Matlab
implementation, nearly $15$ times faster than the LL classifier.

\subsection{Heterogeneous Particle Sizes}
\label{subsec:PVdata}

\begin{figure}[!h]
  \subfloat[]{\includegraphics[width=.5\textwidth]{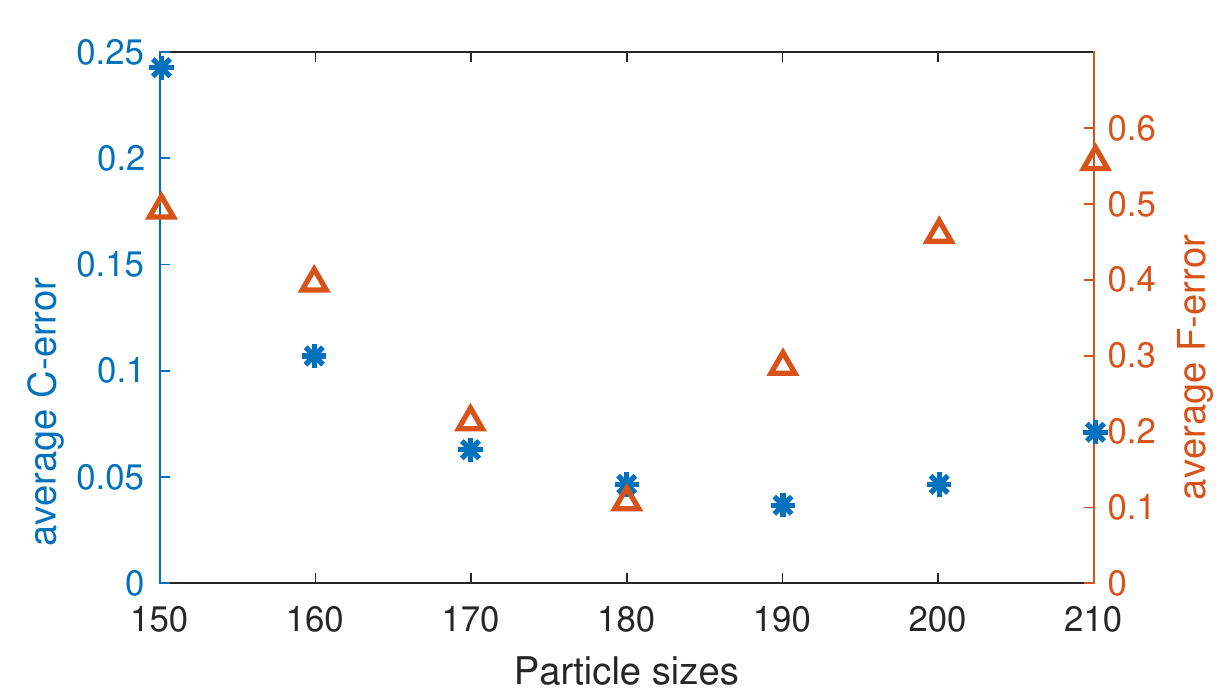} \label{fig:S_cf_eigen}}
  \subfloat[]{\includegraphics[width=.5\textwidth]{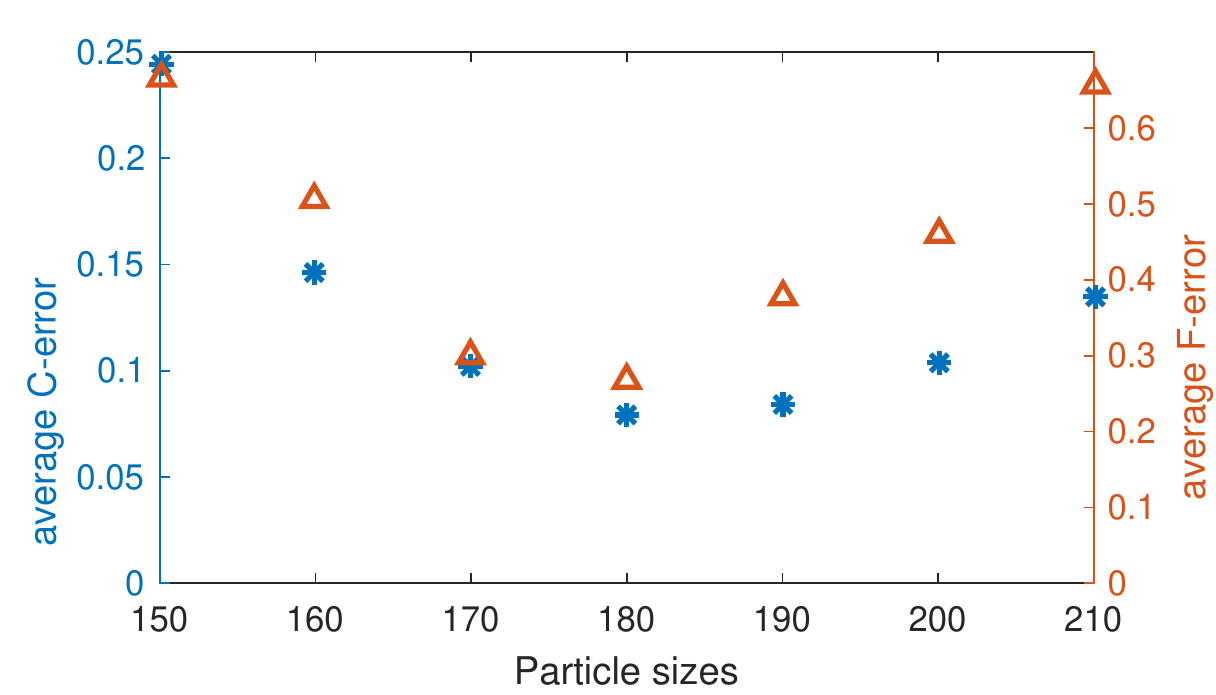}\label{fig:S_cf_ll}}\\
  \subfloat[]{\includegraphics[width=.5\textwidth]{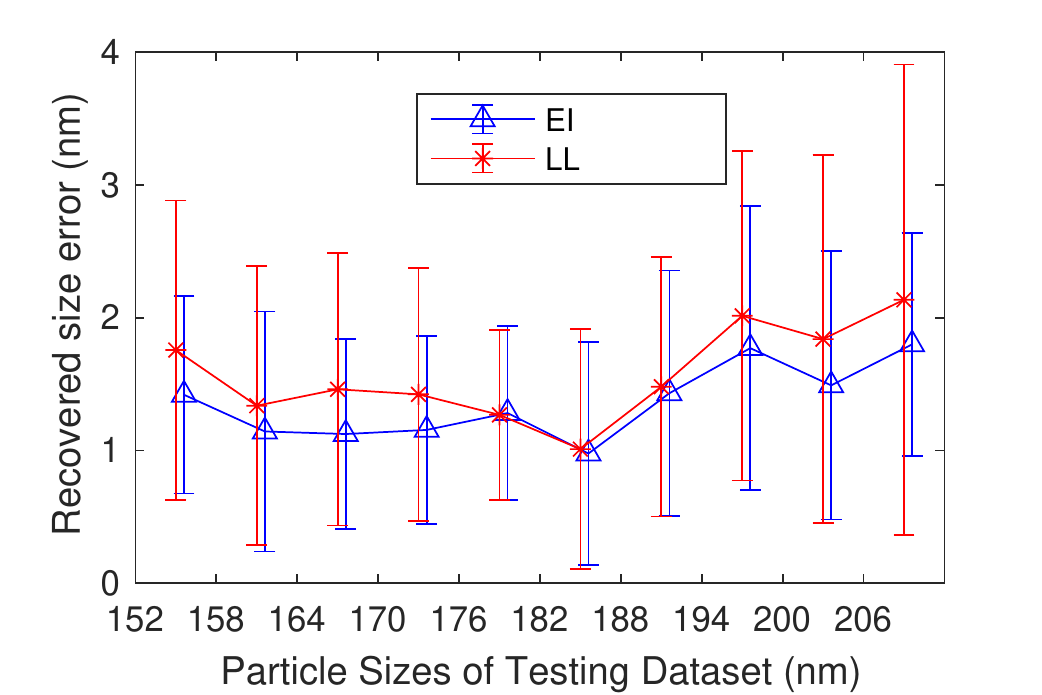} \label{fig:S_cfS}}
  \caption{The classification error (C-error) and the fluence error
    (F-error) of dataset $S$ from the EI
    \protect\subref{fig:S_cf_eigen} and the LL
    \protect\subref{fig:S_cf_ll} classifier. The classification and
    the fluence error were defined in \eqref{eq:L2_S} and
    \eqref{eq:L2_phi_error}, respectively. Both classifiers obtained
    the smallest errors around the template size ($180$
    $nm$). \protect\subref{fig:S_cfS}: The absolute errors in
    estimating sizes from the EI (\emph{blue triangle}) and the LL
    (\emph{red star}) classifier. On average we obtained a minimum
    error of 1~$nm$ around 180~$nm$, and a maximum error of 4 $nm$.  }
  \label{fig:S_cf}
\end{figure}

We next classified the dataset $S$, which contained patterns from
icosahedrons of diameter 150 $nm$ to 210 $nm$. Figure~\ref{fig:S_cf}
illustrates the average classification and fluence errors for the EI
and the LL classifiers.  As expected, both classifiers obtained the
smallest errors when the particle size of the testing pattern was
similar to the template size (180 $nm$), and the LL classifier had
slightly larger errors on average. Furthermore, the EI classifier was
better at estimating particle sizes, as shown in
Figure~\protect\subref*{fig:S_cfS}, and this also implied that the EI
classifier was more accurate in searching for the best-matched
template than the LL classifier was.

The size and the fluence estimation procedures together took around
$80$ $ms$ for each image, i.e.,~around 1.5 times longer than the LL
classifier or 22 times longer than the EI classifier. In other words,
with size estimation, the EI classifier can handle 12 images per
minute and the LL classifier can perform 8 images per minute, as using
a single-core Matlab implementation. With 16 cores, it is therefore
possible to speed up both classifiers to the LCLS repetition rate (120
Hz). Using a larger numbers of GPUs and/or CPU cores, we may also
parallelize them to match the European XFEL rate (2,700 Hz).

\subsection{Heterogeneous Shapes}
\label{subsec:Mdata}

\begin{figure}[!htbp]
  \subfloat[]{\includegraphics[width=.5\textwidth]{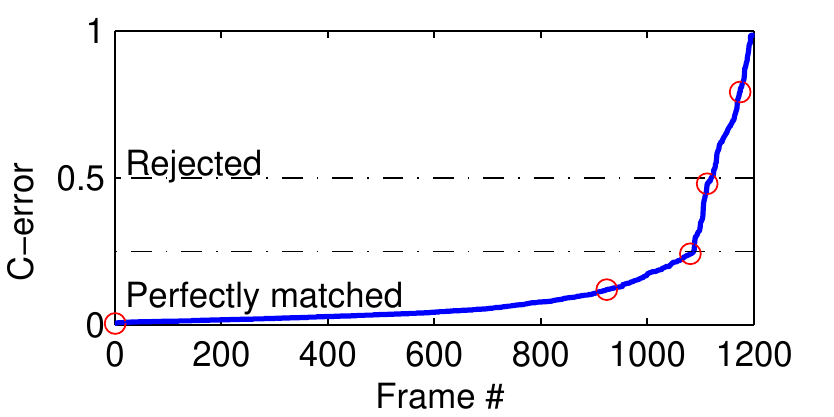} \label{fig:M_f_cerr}}
  \subfloat[]{\includegraphics[width=.5\textwidth]{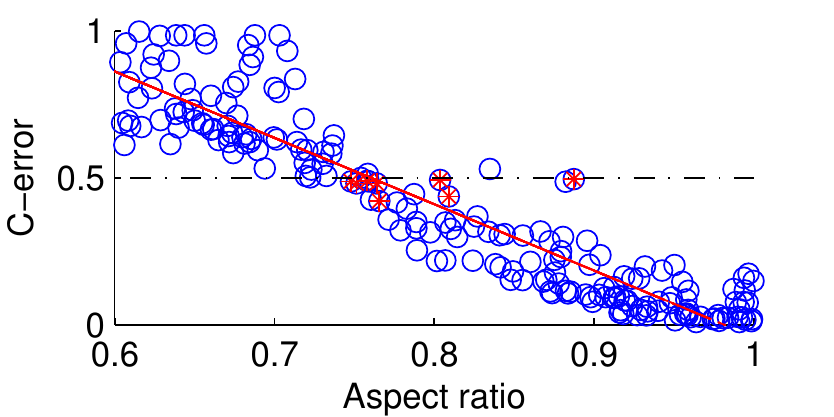} \label{fig:M_f_ratio}}\\
  \subfloat[]{\includegraphics[width=.16\textwidth]{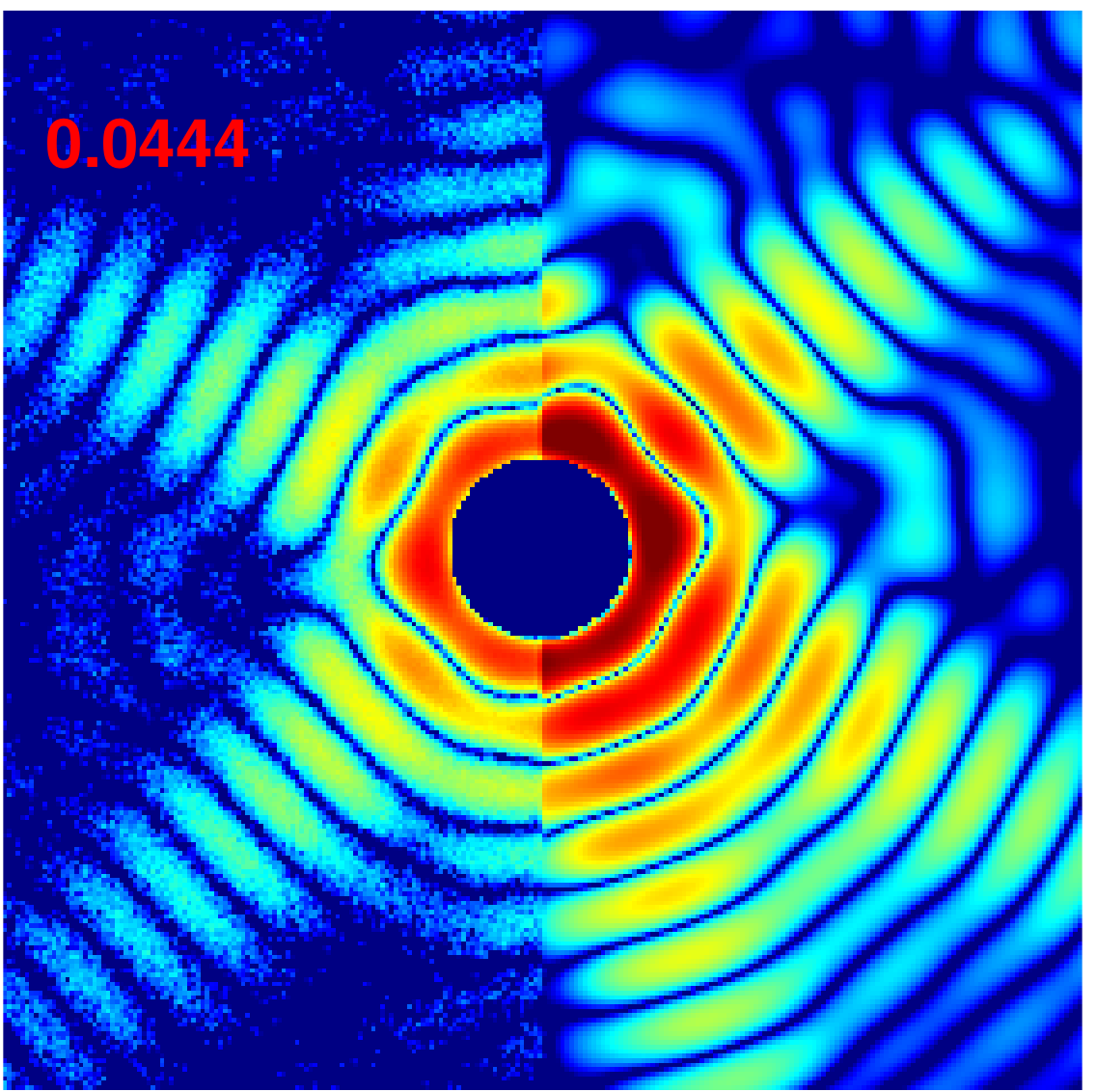} \label{fig:M_f_c}}
  \subfloat[]{\includegraphics[width=.16\textwidth]{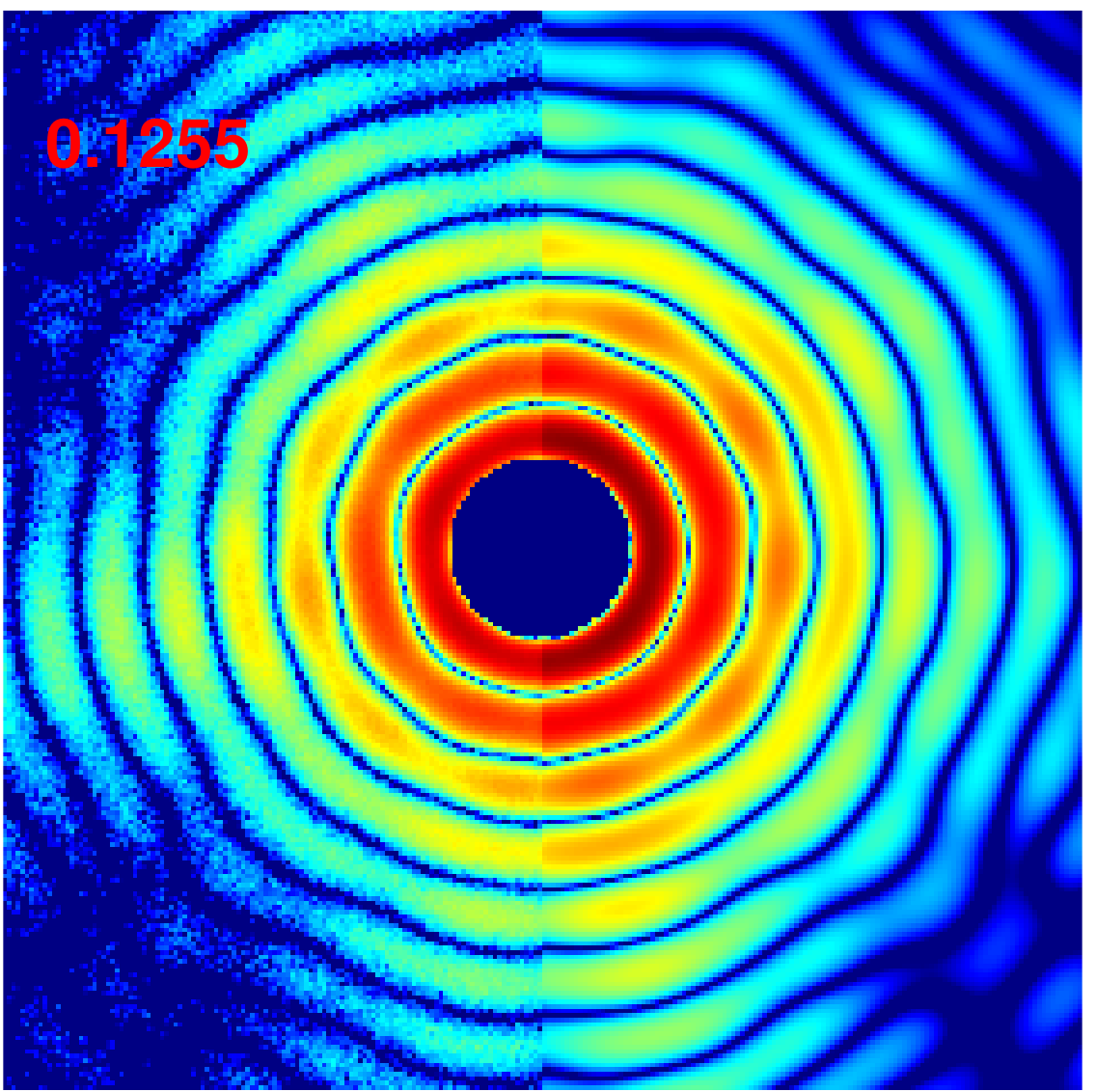}}
  \subfloat[]{\includegraphics[width=.16\textwidth]{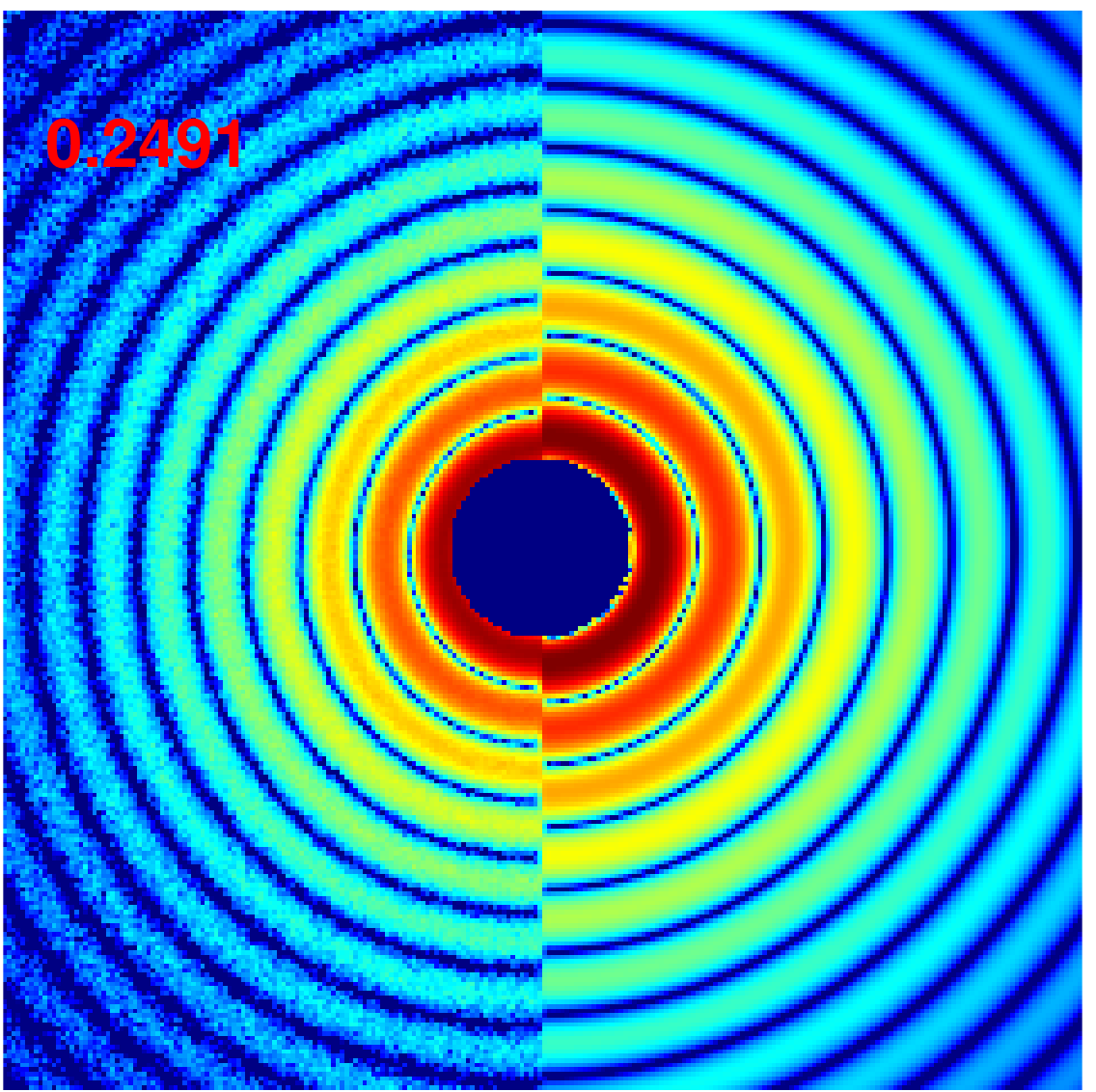}}
  \subfloat[]{\includegraphics[width=.16\textwidth]{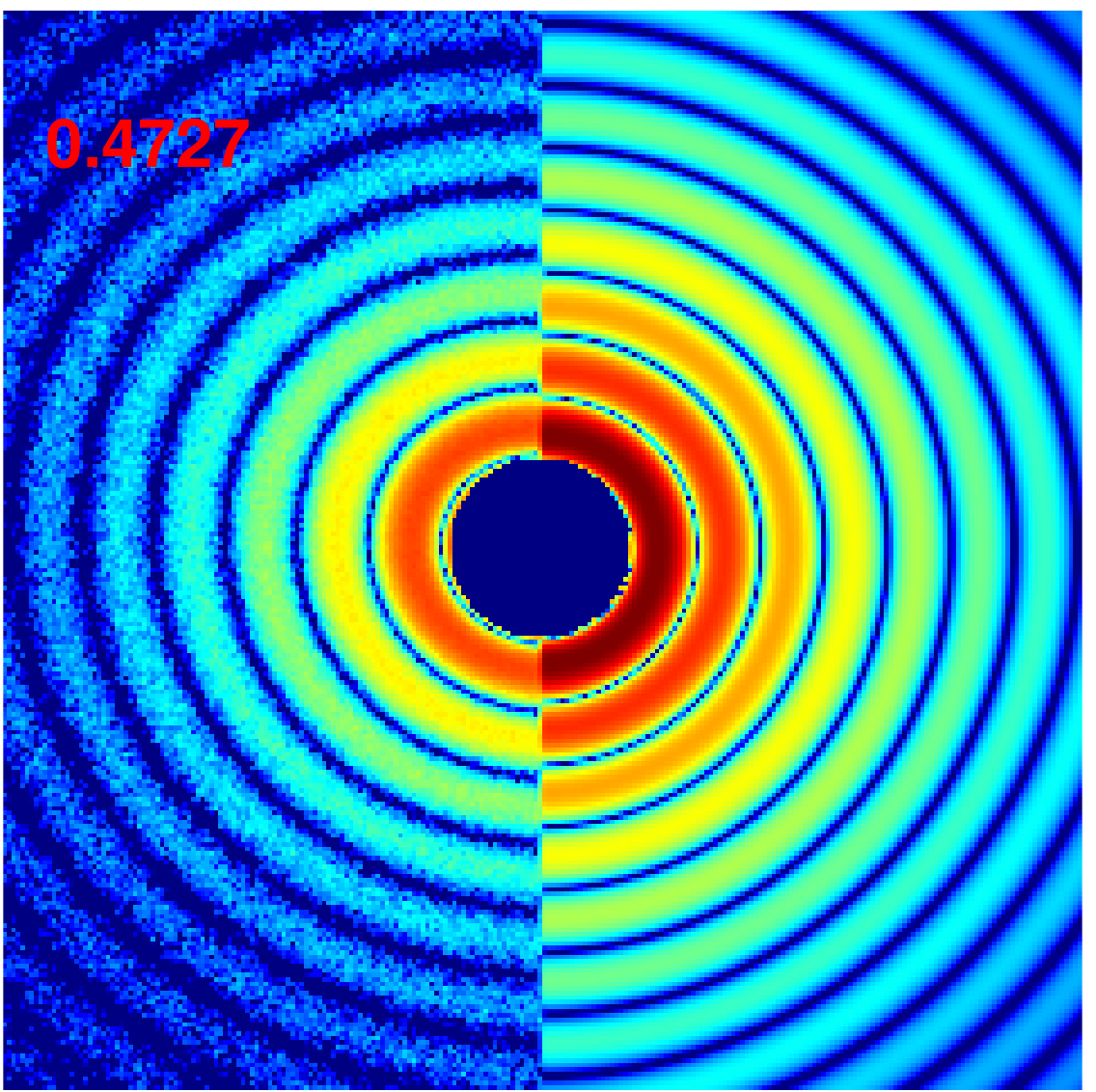}}
  \subfloat[]{\includegraphics[width=.16\textwidth]{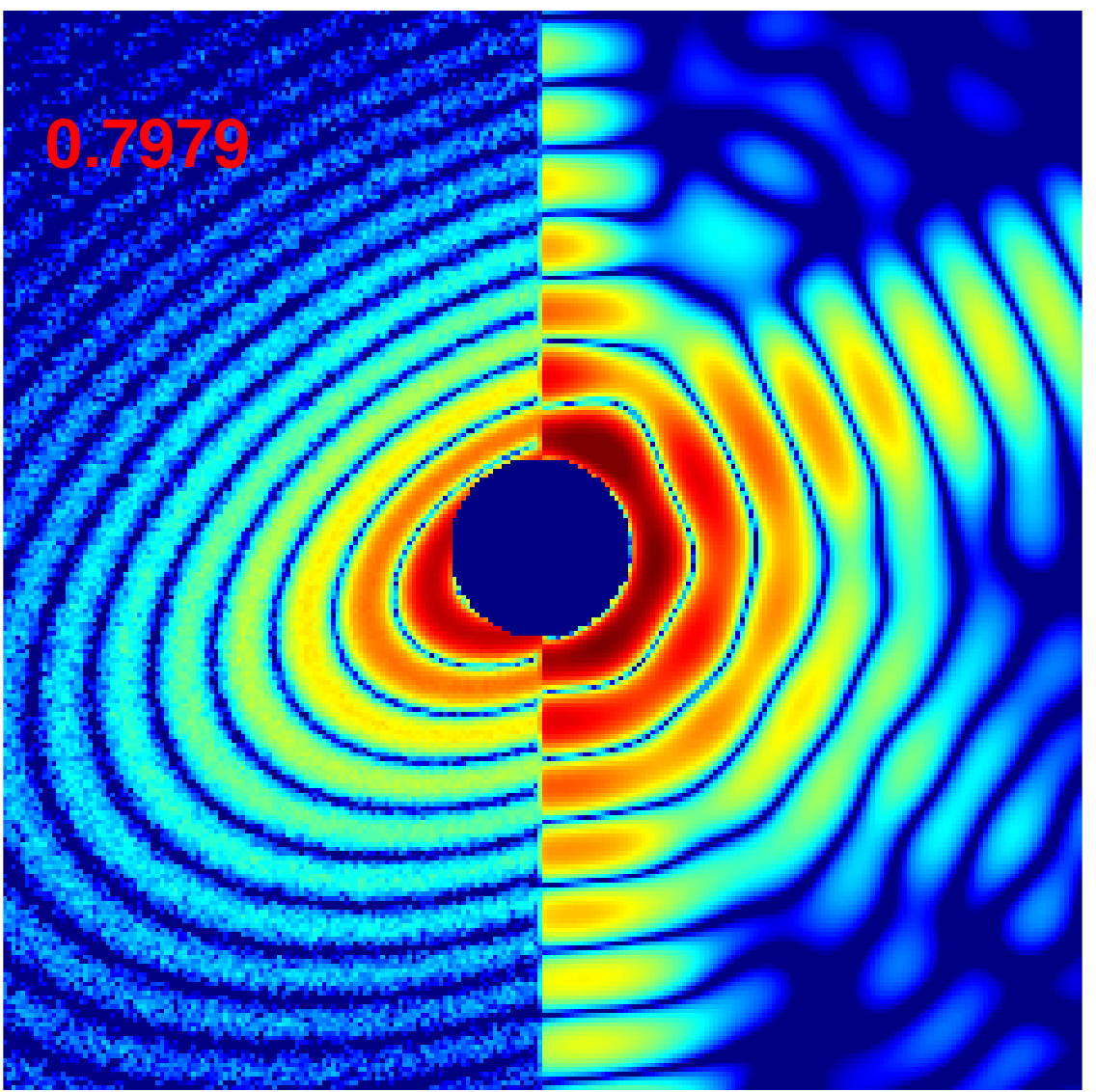} \label{fig:M_f_g}}\\
  \caption{\protect\subref{fig:M_f_cerr}: The classification errors
    (C-error) of dataset $X$.  All icosahedral patterns were located
    in the perfectly matched region, and all elongated spheroidal
    patterns (78 patterns) were rejected. For the rest $122$ accepted
    spheroidal patterns, 114 were successfully classified as
    spheroids, and 8 frames, or 4\% of the spheroidal patterns were
    misclassifed. \protect\subref{fig:M_f_ratio}: The relationship
    between the C-error and the aspect ratios for the spheroidal
    patterns. The red stars were misclassified patterns.  The aspect
    ratio was the ratio of the length of the minor axis to the length
    of the main axis of the spheroidal particle.
    [\protect\subref{fig:M_f_c}--\protect\subref{fig:M_f_g}]: Five
    combination images, corresponding to the five data points
    (\textit{red circles}) in \protect\subref{fig:M_f_cerr}. The left
    half of each image was from the testing dataset $X$ and the right
    half was the best-matched patterns from the training dataset. The
    number in each figure was the classification error.}
  \label{fig:M_f}
\end{figure}

Since the EI classifier performed better than the LL classifier in the
previous experiments, we investigated its performance for the dataset
$X$, which contained particles with heterogeneous shapes and
sizes. For identifying the spheroids in $X$, we added a 180~$nm$
sphere diffraction pattern into the training dataset, and retrained
the EI classifier.  The new EI classifier distinguished the
icosahedral and spheroidal diffraction patterns successfully, as
listed in Table~\ref{tab:M_f}.  All icosahedral diffraction patterns
were classified as icosahedron with a smaller classification error
($<0.25$). With a threshold of 0.5, the retrained classifier rejected
78 elongated spheroidal patterns, and identified 114 spheroidal frames
as spheroids successfully. However, 8 (4\%) frames were misclassified
as icosahedron, and their classification errors were between 0.42 and
0.5, see Figure~\subref*{fig:M_f_ratio}.

\begin{table}[!htbp]
  \centering
  \caption{Classification results of dataset $X$.  The threshold of the classification error for rejection was set to 0.5.}
  \begin{tabular}{l |llll}
    \hline
    & \multicolumn{4}{c}{Classified as}                                                                                \\ 
    \multicolumn{1}{l|}{} & \multicolumn{1}{c}{Icosahedron} &\multicolumn{1}{c}{Spheroid} & \multicolumn{1}{c}{Rejected} & \multicolumn{1}{c}{Total} \\ \hline
    Data: icosahedron         & 1000                  & 0           & 0        & 1000            \\
    Data: spheroid            & 8                   & 114  & 78             & 200                \\ \hline
  \end{tabular}
  \label{tab:M_f}
\end{table}
We visually illustrate the classification results in
Figure~\ref{fig:M_f}. With a classification-error threshold of 0.25,
the EI classifier identified all the icosahedral diffraction patterns
and 86 roundish spheroidal patterns. With a threshold of 0.5, the
classifier rejected 78 elongated spheroidal patterns, and most aspect
ratios of these rejected patterns were less than 0.75, see
Figure~\subref*{fig:M_f_ratio}. As expected, we also observed that the
classification error decreased with increasing aspect ratio.

\subsection{Mimivirus diffraction patterns}

\begin{figure}[!htbp]
  \centering
  \subfloat[]{\includegraphics[width=.25\textwidth]{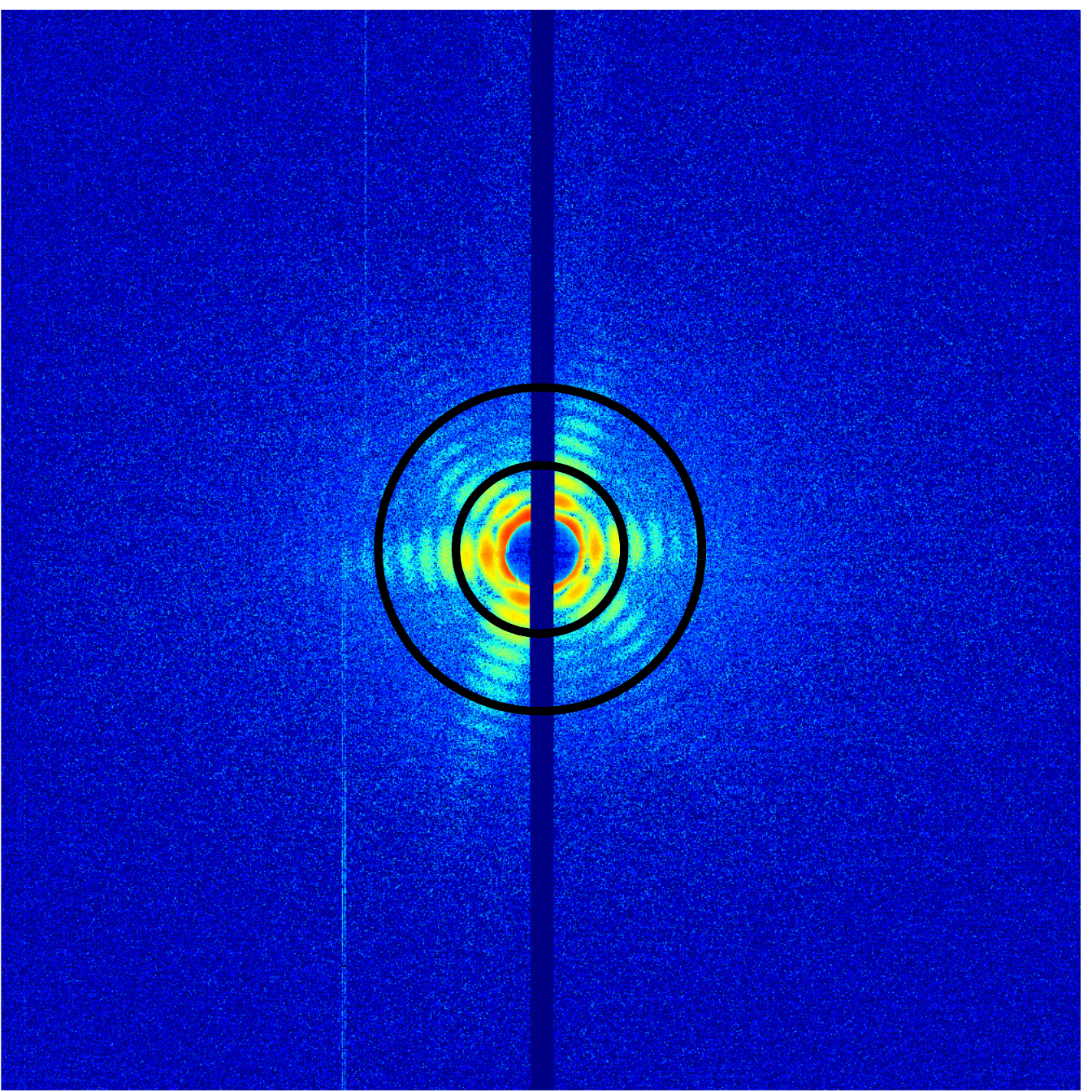}\label{fig:mimi_W}}
  \subfloat[]{\includegraphics[width=.25\textwidth]{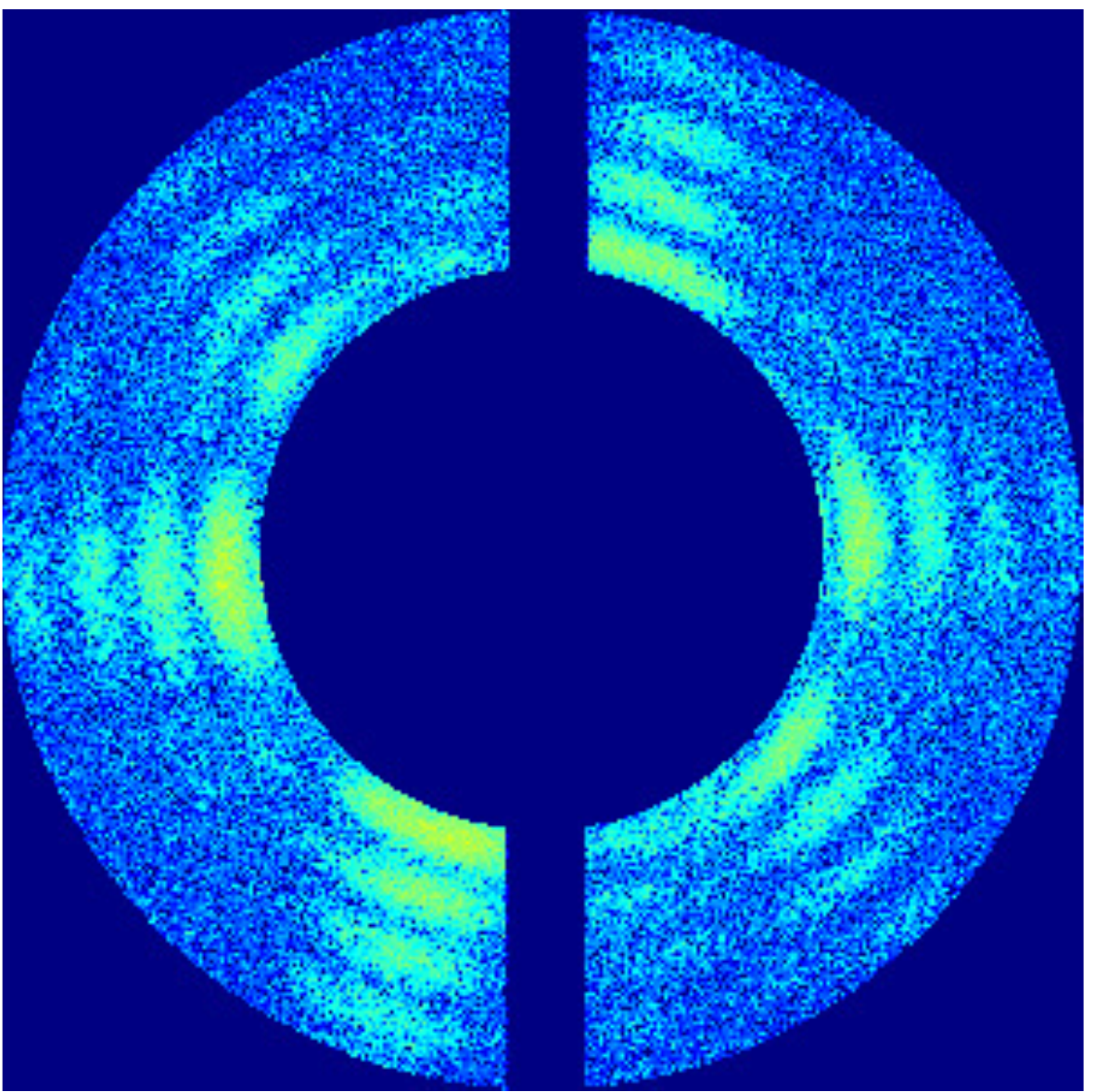}\label{fig:mimi_C}}
  \caption{A mimivirus diffraction pattern \protect\subref{fig:mimi_W}
    and its central region used for classification
    \protect\subref{fig:mimi_C}. The region shown in
    \protect\subref{fig:mimi_C} was the region between two circles in
    \protect\subref{fig:mimi_W}.}
  \label{fig:mimi}
\end{figure}

We finally tested our EI classifier on the mimivirus FXI dataset,
which has been used for 3D mimivirus reconstruction \cite{3d_mimi}. To
compensate detector saturation at the image center and low
signal-to-noise ratio at the edges of the patterns, we used the
central part of the diffraction patterns for classification, see
Figure~\ref{fig:mimi}. The training dataset for the mimivirus dataset
contained 1000 randomly-oriented icosahedral patterns of $490$ $nm$ in
diameter. Furthermore, we binned $4\times 4$ pixels into one pixel in
the calculations.

\begin{figure}[!htbp]
  \centering
  \subfloat[]{\includegraphics[width=.5\textwidth]{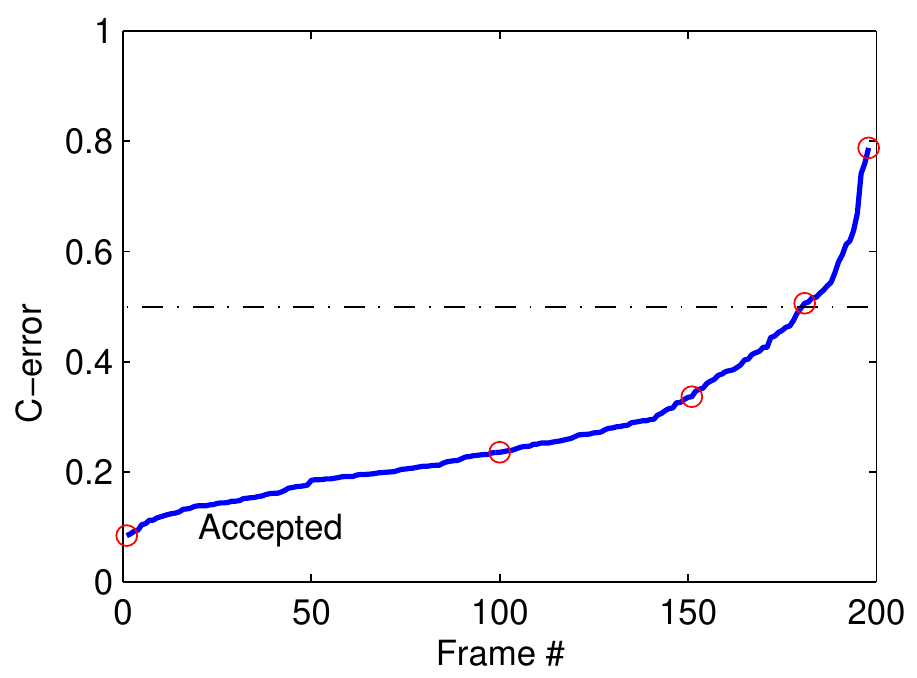}\label{fig:mimi_r_err}} 
  \subfloat[]{\includegraphics[width=.5\textwidth]{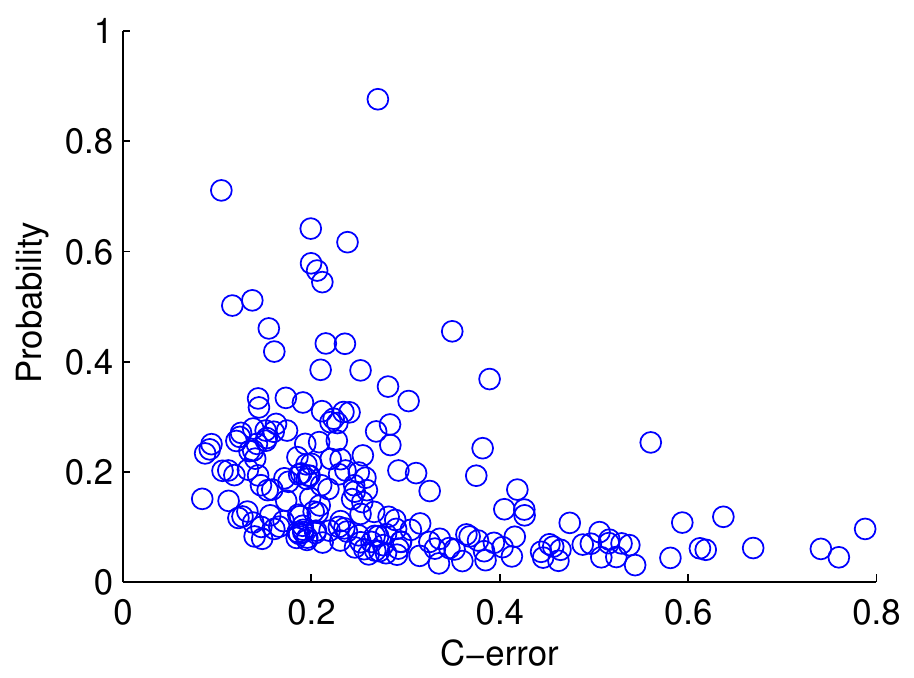}\label{fig:mimi_validate3D}} 
  \\
  \subfloat[]{\includegraphics[width=.2\textwidth]{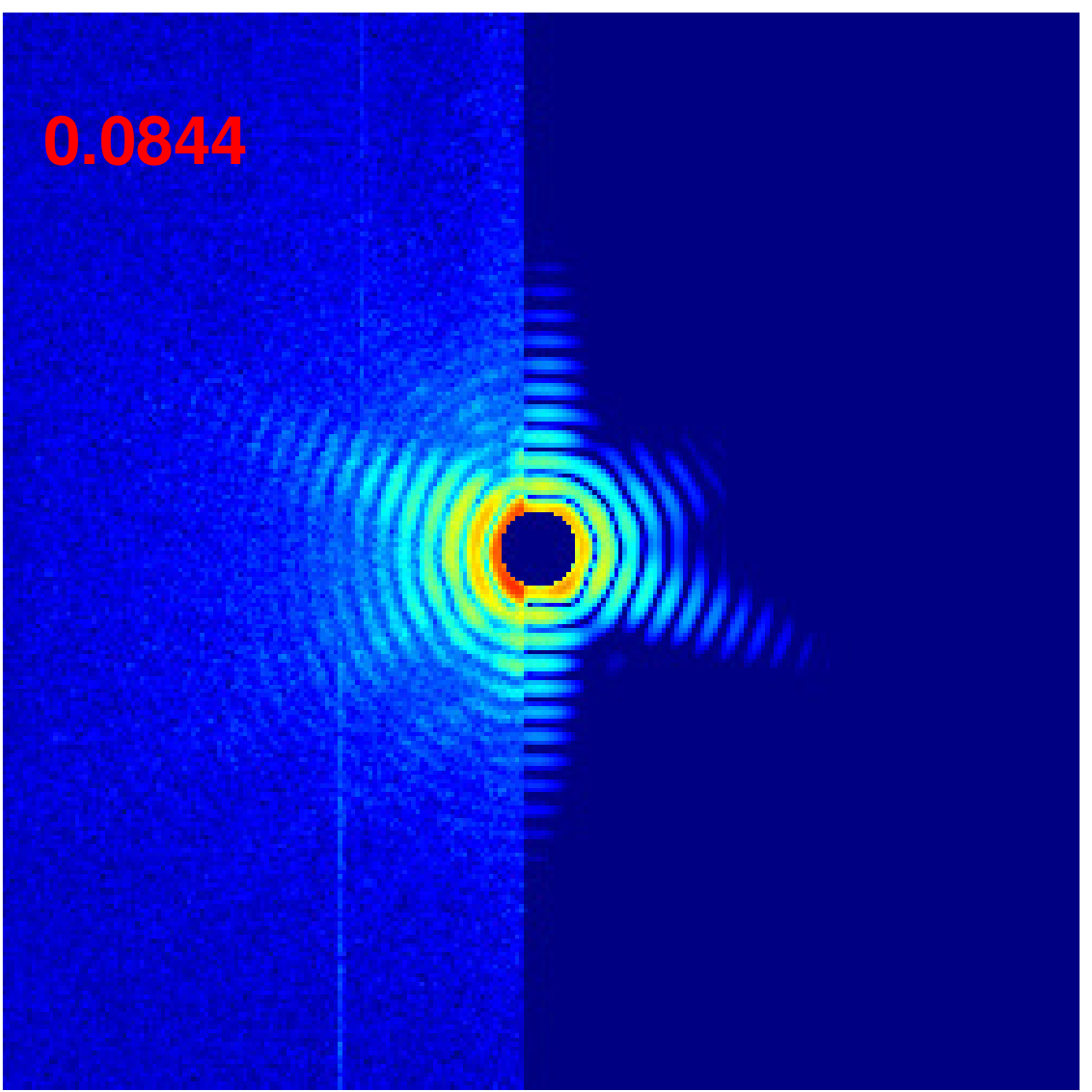}\label{fig:mimi_r_1}}
  \subfloat[]{\includegraphics[width=.2\textwidth]{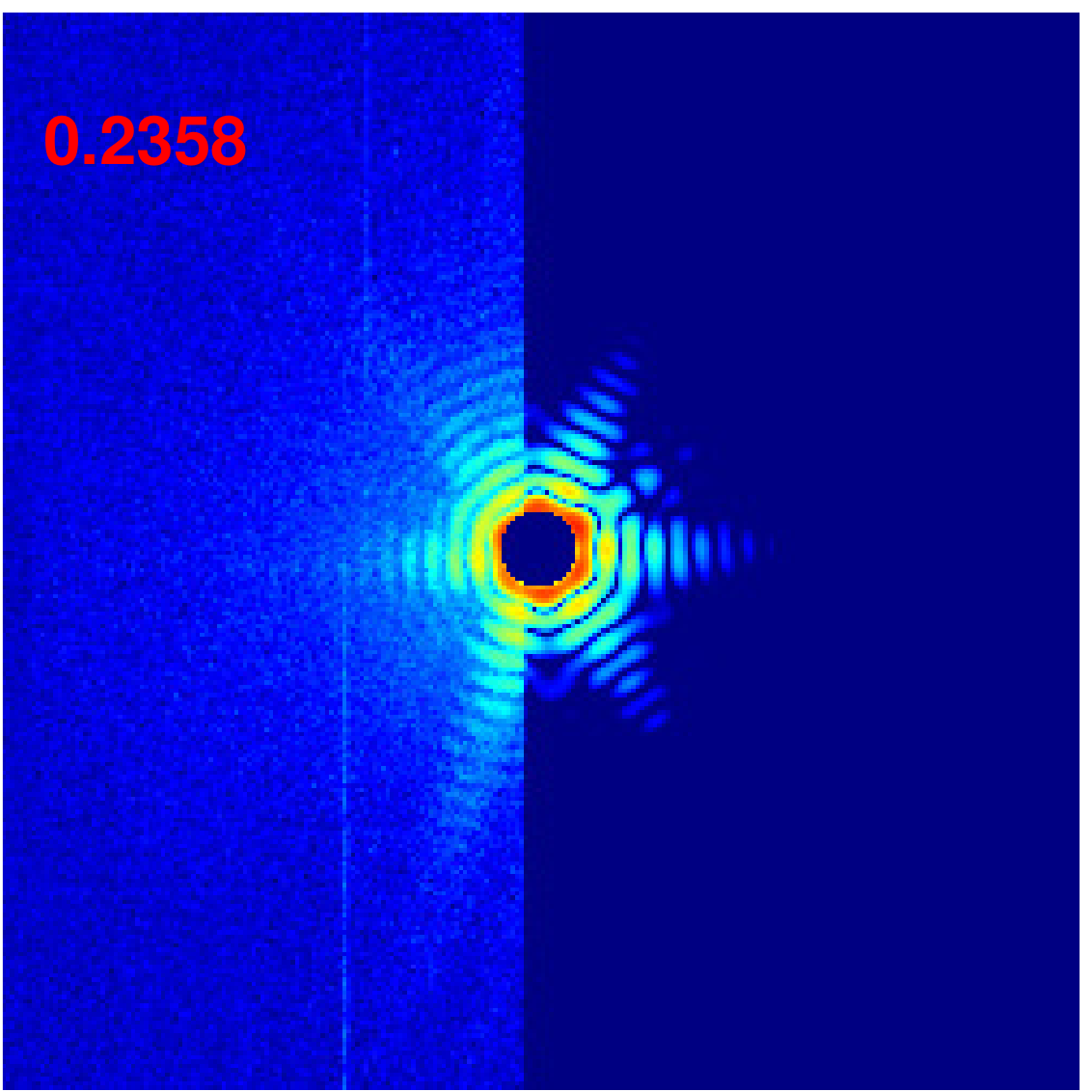}\label{fig:mimi_r_2}}
  \subfloat[]{\includegraphics[width=.2\textwidth]{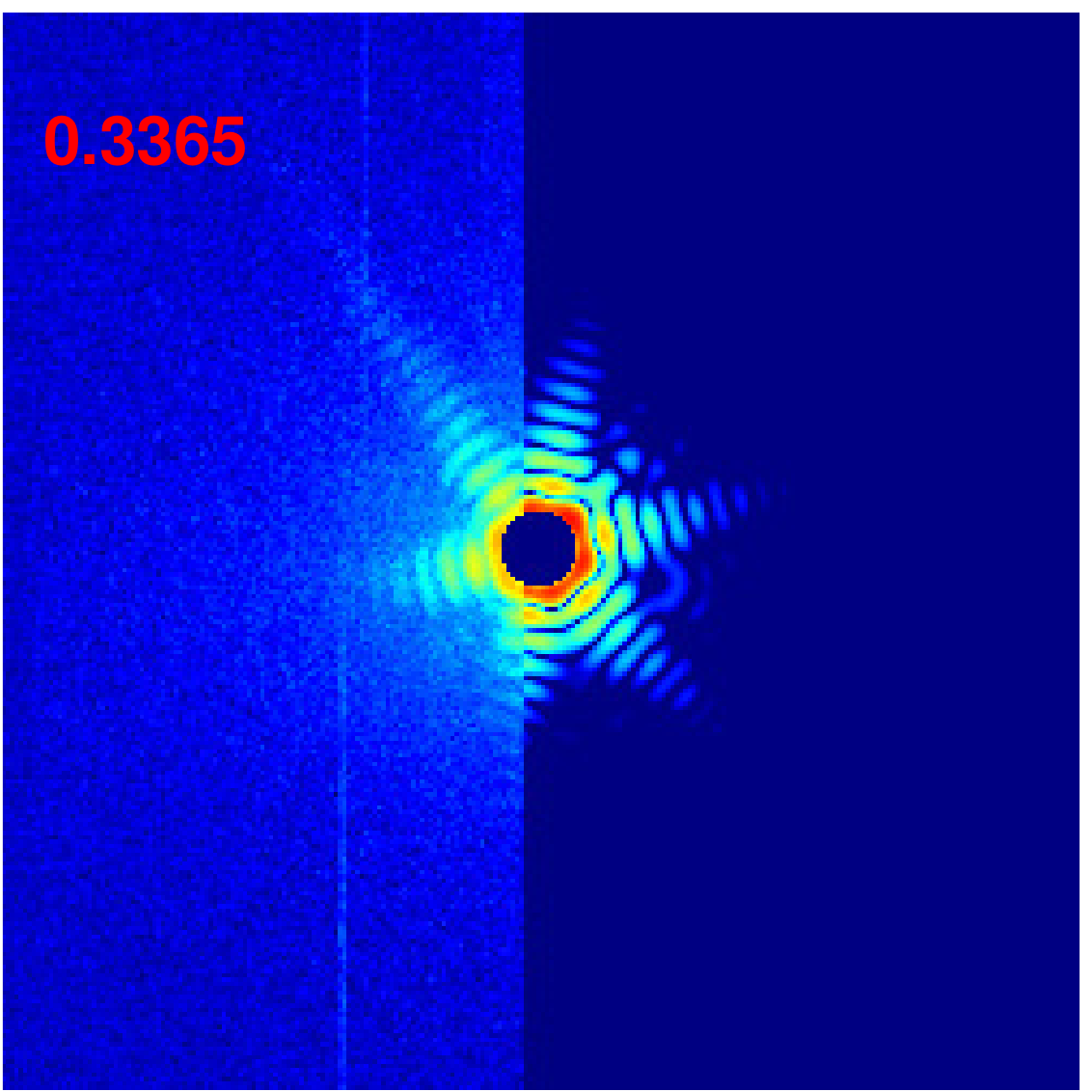}\label{fig:mimi_r_3}}
  \subfloat[]{\includegraphics[width=.2\textwidth]{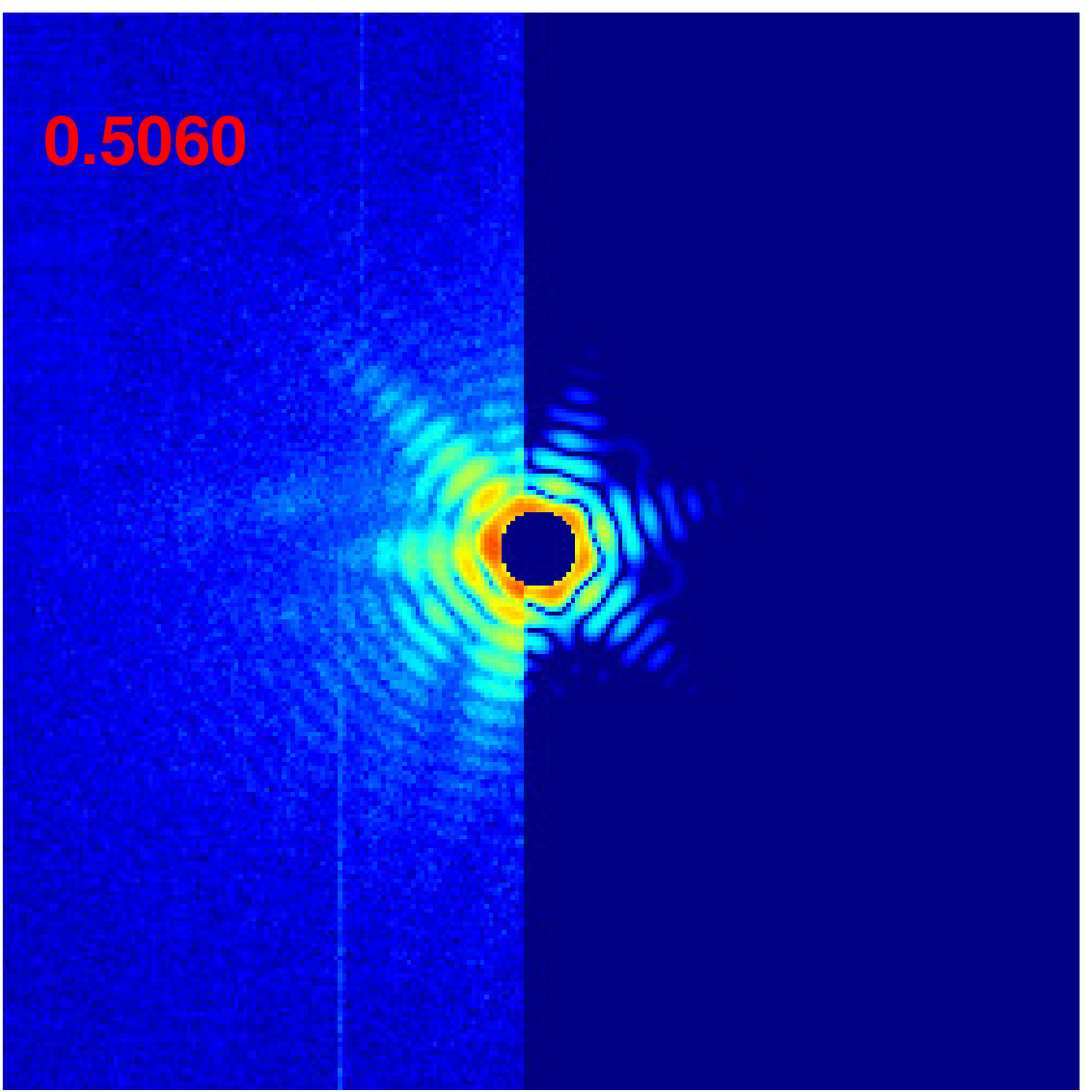}\label{fig:mimi_r_4}}
  \subfloat[]{\includegraphics[width=.2\textwidth]{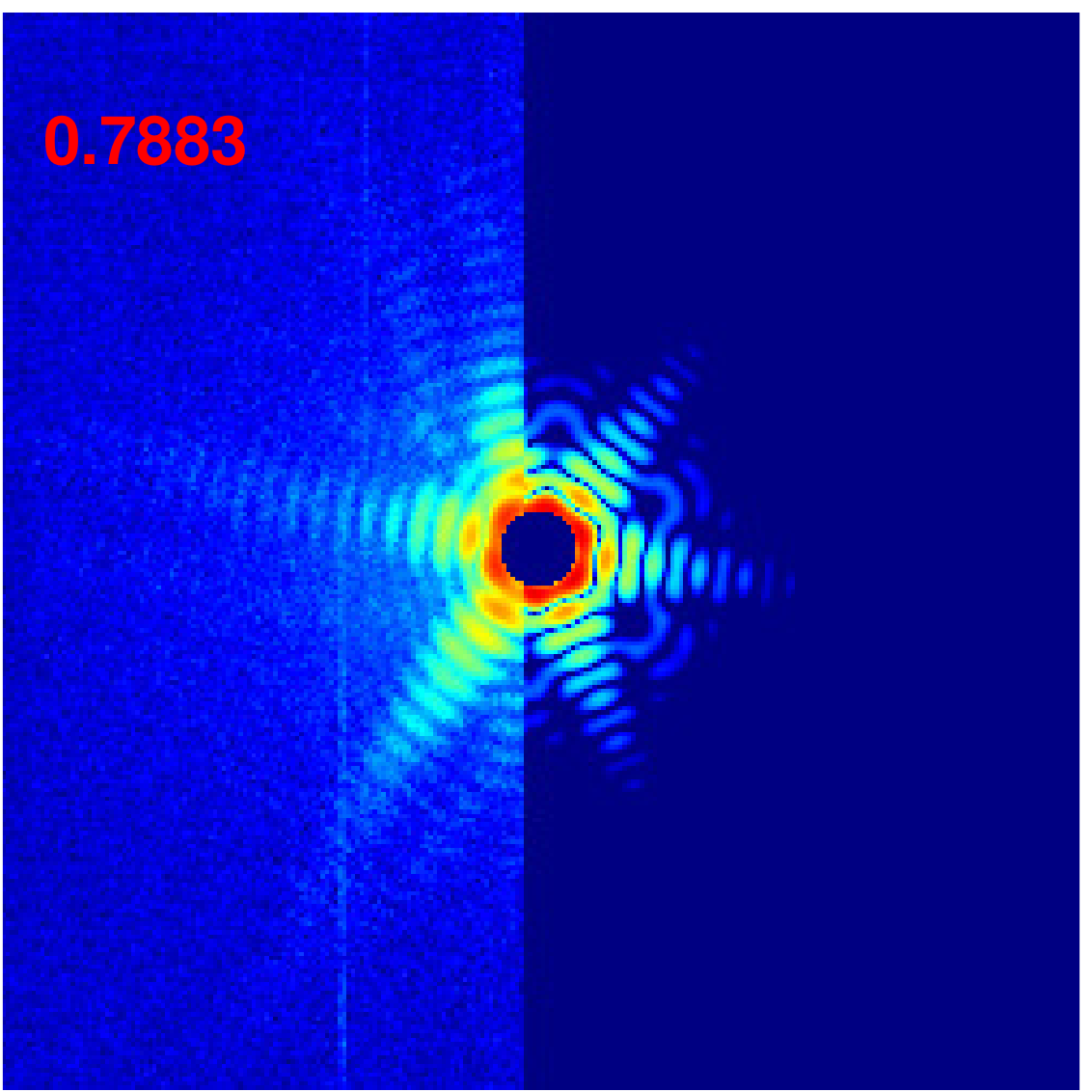}\label{fig:mimi_r_5}}
  \caption{\protect\subref{fig:mimi_r_err}: The classification error
    (C-error) of the mimivirus dataset from the EI Classifier. 16 of
    198 patterns (9.1\%) were rejected with a threshold of
    0.5. \protect\subref{fig:mimi_validate3D}: The relationship
    between the C-error and the sum of the largest 0.035\% rotational
    probabilities of each diffraction
    pattern. [\protect\subref{fig:mimi_r_1}--\protect\subref{fig:mimi_r_5}]:
    Five combination images at the data points (\textit{red circles})
    in \protect\subref{fig:mimi_r_err}.  The \emph{left} part of each
    image was from the mimivirus dataset, and the \emph{right} part
    was the corresponding template scaled by the recovered fluence.
    \protect\subref{fig:mimi_r_4} was a slightly elongated pattern and
    the particle size of \protect\subref{fig:mimi_r_5} was smaller
    than the template size.  }
\label{fig:mimi_r}
\end{figure}

As expected, we obtained larger classification errors, see
Figure~\ref{fig:mimi_r}, and this is due to the heterogeneity in size
and shape of the mimiviruses. We used 0.5 as a threshold to detect
irregular patterns. In total, 9.1\% of patterns (18 patterns) were
rejected. We also validated our classification results to the 3D
Fourier intensity by looking at the correlation between the
classification errors and the sum of the largest 0.035\% rotational
probabilities of each diffraction pattern in
Figure~\subref*{fig:mimi_validate3D}. We assembled the 3D Fourier
intensity by the Expansion Maximization Compression (EMC) method with
an assumption of scaled Poisson noise \cite{algEMC}.  As expected, the
sum of the rotational probabilities increased with decreasing
classification error. However, we did not get a linear correlation,
most likely due to the fact that the mimiviruses samples were not
regular uniform-density icosahedrons, and had different particle
sizes. For example, in
[Figure~\subref*{fig:mimi_r_1}--Figure~\subref*{fig:mimi_r_3}], the
templates and the mimivirus patterns matched quite well, however, the
particle in Figure~\subref*{fig:mimi_r_4} was slightly elongated, and
the particle size of Figure~\subref*{fig:mimi_r_5} was 457 $nm$, which
was 33 $nm$ smaller than the template size (490 $nm$).

%**************************************************************************

\section{Conclusions}
\label{sec:Conclusion}

% recall to the XFEL and FXI and difficulties
The FXI technique holds the promise of obtaining biomolecule
structures from single particles. It operates at a high repetition
rate and records thousands of millions of diffraction data every
day. The stochastic nature of XFELs and the heterogeneity of the
sample molecules make the recorded dataset too complex and massive to
classify manually. By using our knowledge of the sample molecules,
such as sizes and shapes, we can use template-based methods to reduce
the complexity of the classification problem.

% recall results and discuss
To find the best-matched pattern, we have presented and tested two
supervised learning methods --- the Eigen-Image (EI) and the
Log-Likelihood classifier. In our straight-forward Matlab
implementations, both methods can classify a testing pattern in a few
milliseconds, and they certainly can be accelerated to the XFEL
repetition rates, albeit using considerable resources.  We also
observed that the rotational probabilities from the 3D assembling
procedure, increased with decreasing classification error. This
suggests that the selected patterns from our classifiers will fit
better into a 3D Fourier intensity, resulting in a potentially
high-resolution 3D electron density of the sample molecules.

%future work
Newer facilities, such as the European XFEL, operate at high
repetition rates and will create massive volumes of FXI diffraction
data with heterogeneities to varying degrees. With our methods, we can
use most of our knowledge of the sample molecules to reduce data
storage and automatically select homogeneous single-particle
patterns. We also foresee that an on-site FXI analysis pipeline, which
connects our classifier to the 3D reconstruction procedure, can solve
the 3D structure with sub-nanometer resolution in the near future.

%**************************************************************************

\section*{Acknowledgment}
	
This work was financially supported  by the Swedish Research Council
within the UPMARC Linnaeus center of Excellence (S.~Engblom, J.~Liu)
and by the Swedish Research Council, the R\"ontgen-\AA ngstr\" om
Cluster, the Knut och Alice Wallenbergs Stiftelse, the European
Research Council (J.~Liu, G.~Schot).
	
%**************************************************************************

\newcommand{\doi}[1]{\href{http://dx.doi.org/#1}{doi:#1}}
\newcommand{\available}[1]{Available at \url{#1}}
\newcommand{\availablet}[2]{Available at \href{#1}{#2}}

\bibliographystyle{abbrvnat}
\bibliography{Psort}
\end{document}